%% file: templateArxiv.tex
\documentclass{article}

\usepackage{PRIMEarxiv}

\usepackage[utf8]{inputenc} 
\usepackage[T1]{fontenc}    
\usepackage{hyperref}       
\usepackage{url}            
\usepackage{booktabs}       
\usepackage{amsfonts}       
\usepackage{nicefrac}       
\usepackage{microtype}      
\usepackage{lipsum}
\usepackage{fancyhdr}       
\usepackage{graphicx}       
\graphicspath{{media/}}     

\usepackage{algorithm}
\usepackage{algpseudocode}
\usepackage{amsmath,amssymb}
\usepackage{multirow}
\usepackage{booktabs}
\usepackage{array}
\usepackage{pifont}
\newcommand{\cmark}{\ding{51}}
\usepackage{booktabs}

\title{InvZW: Invariant Feature Learning via
Noise-Adversarial Training for Robust Image
Zero-Watermarking
}

\author{
  Abdullah All Tanvir\\
  Department of Computer Science \\
  University of Nebraska at Omaha \\
  Omaha, Nebraska, USA\\
  \texttt{atanvir@unomaha.edu}\\
  \And
  Frank Y. Shih \\
  Department of Computer Science \\
  New Jersey Institute of Technology \\
  Newark, New Jersey, USA\\
  \texttt{shih@njit.edu} \\
  \And
  Xin Zhong\\
  Department of Computer Science \\
  University of Nebraska at Omaha \\
  Omaha, Nebraska, USA\\
  \texttt{xzhong@unomaha.edu}\\
}

\begin{document}
\maketitle

\begin{abstract}
\input{sec/abstract}
\end{abstract}


\input{sec/1_intro}

\input{sec/2_related_work}

\input{sec/3_proposed_method}

\input{sec/4_experimental_result}

\input{sec/5_conclusion}

\bibliographystyle{unsrt}  
\bibliography{references}

\end{document}

%% file: sec/abstract.tex
This paper introduces a novel deep learning framework for robust image zero-watermarking based on distortion-invariant feature learning. As a zero-watermarking scheme, our method leaves the original image unaltered and learns a reference signature through optimization in the feature space. The proposed framework consists of two key modules. In the first module, a feature extractor is trained via noise-adversarial learning to generate representations that are both invariant to distortions and semantically expressive. This is achieved by combining adversarial supervision against a distortion discriminator and a reconstruction constraint to retain image content. In the second module, we design a learning-based multibit zero-watermarking scheme where the trained invariant features are projected onto a set of trainable reference codes optimized to match a target binary message. Extensive experiments on diverse image datasets and a wide range of distortions show that our method achieves state-of-the-art robustness in both feature stability and watermark recovery. Comparative evaluations against existing self-supervised and deep watermarking techniques further highlight the superiority of our framework in generalization and robustness.

%% file: sec/1_intro.tex
\vspace{-1.25em}
\section{Introduction}
\label{sec:intro} 

Image watermarking is a widely used technique for embedding hidden information into digital media for purposes such as copyright protection and content authentication~\cite{zhang2024dual}. 
In contrast to standard watermarking methods that modify the cover image by embedding a watermark directly into its pixel or transform domain representation, zero-watermarking leaves the original image unchanged. 
Instead, it extracts robust features from the image and associates them with a watermark to form a signature, which is stored externally without altering the image. During usage, features extracted from a potentially marked image are used to either verify the watermark or recover it. 
This approach ensures that the visual integrity of the original image remains intact, making it ideal for scenarios where any modification is impermissible. While zero-watermarking has traditionally been associated with highly sensitive use cases, its non-intrusive and robust nature makes it broadly applicable. By enabling reliable copyright protection and content verification without altering the original content, it offers a practical and versatile solution across a wide range of image-based applications. 
There are two primary objectives of zero-watermarking: verification-based approaches, which confirm the presence of a watermark~\cite{li2024zwnet}, and multibit extraction-based methods, which aim to retrieve the full watermark content from the extracted features~\cite{fierro2019robust}. 

\begin{figure}[!th]
    \centering
    \vspace{-1.0em}
    \includegraphics[width=0.4\linewidth]{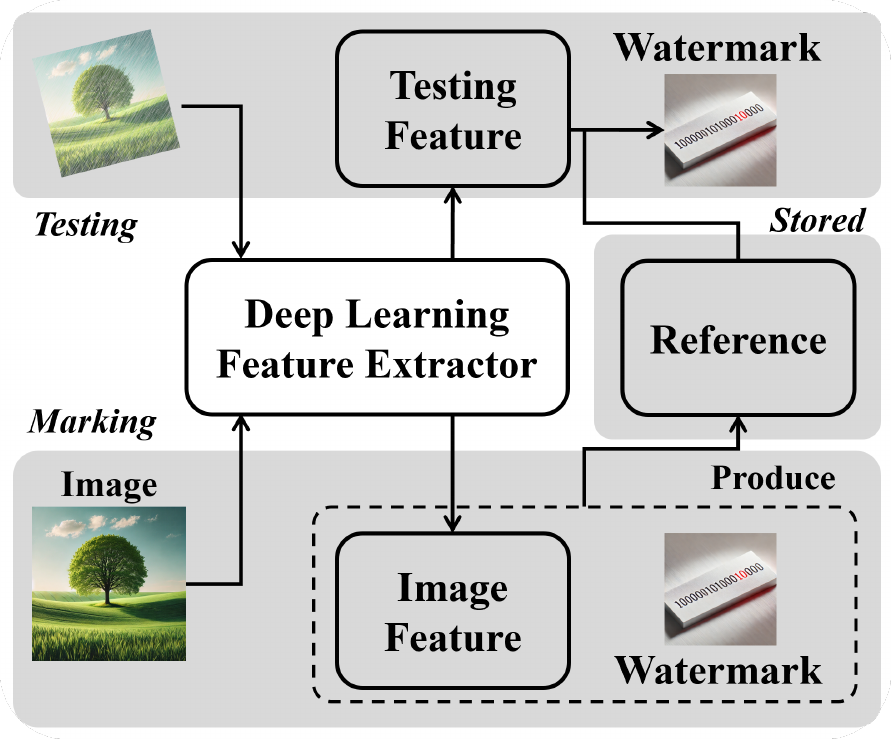}
    \vspace{-0.5em}
    \caption{Image Zero-Watermarking with Deep Learning.}
    \label{fig:DLZW}
    \vspace{-0.75em}
\end{figure}

Recent advancements in deep learning have transformed the landscape of zero-watermarking by introducing learning-based approaches that leverage deep neural networks to extract high-level feature representations from images~\cite{zhong2023brief}. 
Instead of relying on handcrafted descriptors, these methods employ convolutional networks or transformers to learn features that are more robust to distortions~\cite{fierro2019robust,he2023shrinkage,han2024application}. 
A common approach involves using a deep network as a fixed feature extractor: features are drawn from the input image and combined with a watermark to generate a reference signature. During verification, the same feature extraction process is applied to a test image to determine watermark presence or recover watermark bits (see Fig.~\ref{fig:DLZW}). 
Beyond zero-watermarking, several deep image watermarking schemes~\cite{vukotic2020classification, fernandez2022watermarking} also leverage pre-trained networks, especially self-supervised models, as fixed feature extractors. These methods modify the input image to embed a watermark that becomes detectable in the feature space. By inheriting the robustness of self-supervised representations, they offer robustness to distortions without additional training. 
These approaches typically assume that off-the-shelf models, trained on large-scale image classification or self-supervised learning tasks, provide sufficient invariance to potential image distortions in watermarking.

Robustness is a critical consideration in image zero-watermarking, where the extracted features must remain stable under a wide range of distortions. These include common image processing attacks such as blurring and compression, as well as geometric transformations such as rotation, scaling, and cropping. 
These transformations can significantly alter image appearance while preserving semantics, posing a challenge for watermark retrieval. While existing methods have leveraged pre-trained feature extractors to achieve a degree of robustness, they are not explicitly designed for watermarking and often fall short when encountering unseen or complex distortions. This highlights the need for more specialized approaches that directly target invariance in the feature space, tailored to the demands of watermarking applications. 

To this end, we propose a noise-adversarial training framework for learning invariant features specifically designed for multibit image zero-watermarking. 
Our method introduces three core novelties: (1) a new invariant feature learning scheme that is explicitly tailored for watermarking applications, ensuring that feature representations remain stable under diverse distortions; (2) a novel deep learning-based zero-watermarking method that leverages these learned features for watermark extraction without modifying the original image; and (3) a significant improvement in robustness, as our adversarial training approach enables the model to withstand a wide range of image processing and geometric transformations. 
Collectively, these components constitute a framework that advances the robustness and reliability of deep zero-watermarking.

%% file: sec/2_related_work.tex
\vspace{-1.0em}
\section{Related Work}
\label{sec:formatting}

This section reviews the most relevant areas, including invariant feature learning and deep learning-based zero-watermarking.

\subsection{Invariant Feature Learning}

Robust feature learning has its origins in convolutional neural networks (CNNs), which introduce some invariance and equivariance through architectural components such as convolutional filters and pooling operations~\cite{lecun1998cnn}. 
Recent advancements have shifted towards unsupervised and data-driven approaches for determining robustness, reflecting a significant change in paradigm~\cite{autoaugment2018, zhang2021learning, kong2023understanding}. A key development is the use of autonomous invariance learning through methods like differentiable Kronecker-factored Laplace approximations, which eliminate the need for manual data augmentation selection by optimizing augmentation parameters directly during training~\cite{immer2021invariance}. This enables models to learn transformation robustness from data, adapting to varied conditions without predefined assumptions. 
While Bayesian approaches to invariance learning have been successfully applied in simpler models like Gaussian processes~\cite{bayesian2018}, extending these methods to deep networks remains a challenge. 
Some advances in scalable marginal likelihood approximations, such as those using the structured Gauss-Newton method, have made it feasible to apply these techniques in deep learning~\cite{laplace1992, kfac2015}. This evolution marks a shift from predefined augmentation strategies towards adaptive, data-driven methods, which align closely with the goals of robust feature extraction.

Self-supervised learning (SSL) methods, such as SimCLR~\cite{chen2020simple}, BYOL~\cite{grill2020bootstrap}, and DINO~\cite{caron2021emerging}, contribute to this trend by implicitly learning invariance through contrastive objectives or teacher-student frameworks during training. However, these methods primarily focus on improving the quality of learned representations, not explicitly on enhancing robustness against a wide range of distortions. Some work has attempted to address this gap in graph neural networks through methods, which employs tailored objectives to disentangle invariant features from spurious correlations~\cite{yao2024empowering}.

Our approach extends these principles by directly training a deep learning model for invariance, with a specific focus on enhancing robustness. Unlike SSL approaches that treat invariance as a byproduct, our method is designed to optimize feature stability under diverse image transformations. This makes it particularly suitable for applications of image watermarking, where robustness to distortions is critical.

\vspace{-0.75em}
\subsection{Deep Learning-based Image Zero-Watermarking}
\vspace{-0.25em}

A common strategy behind recent deep learning-based image zero-watermarking methods is to adopt an existing deep model, typically pre-trained on large-scale datasets, as a feature extractor and assume that the learned features offer sufficient robustness for watermarking tasks. 
Fierro et al.~\cite{fierro2019robust} utilized a CNN to extract features from the cover host image, which were then combined with a permuted binary watermark using an XOR operation to generate a master share for later verification. 
He et al.~\cite{he2023shrinkage} extended this pipeline by introducing fully connected layers to extract shallow features from multiple convolutional layers, while incorporating a shrinkage module for soft thresholding and a noise layer during training to enhance robustness. 
Han et al.~\cite{han2024application} further improved this approach by encrypting the watermark using a chaotic encryption scheme and employing the Swin Transformer~\cite{liu2021swin} as the feature extractor to improve geometric invariance and overall robustness.
Li et al.~\cite{li2024zwnet} proposed a zero-watermarking framework that uses ConvNeXt and LK-PAN to extract local and global features, with a dedicated watermark block designed to enhance both robustness and discriminability without handcrafted features. 

Beyond zero-watermarking, several deep image watermarking schemes~\cite{vukotic2020classification, fernandez2022watermarking} have explored the use of pre-trained networks, particularly self-supervised models—as fixed feature extractors. These methods embed watermarks by optimizing the input image while keeping the model weights frozen, allowing the watermark to be revealed in the feature space. By leveraging the inherent robustness of self-supervised representations, these approaches achieve resilience to common distortions without the need for retraining the feature extractor.

Current approaches typically assume that off-the-shelf models, trained for image classification or self-supervised tasks, offer sufficient invariance for watermarking. In contrast, we train a deep network specifically to learn invariant features tailored for watermarking, resulting in greater robustness across diverse transformations. 
Moreover, unlike many existing methods that rely on manually computing feature-watermark correlations for verification, our approach introduces a novel learning-based mechanism that derives the reference signature from the extracted invariant features and the target watermark, enabling more reliable and adaptive zero-watermarking.

%% file: sec/3_proposed_method.tex
\vspace{-0.5em}
\section{Proposed Method}
\label{sec:methods}

\begin{figure}[!ht]
    \centering
    \includegraphics[width=0.7\linewidth]{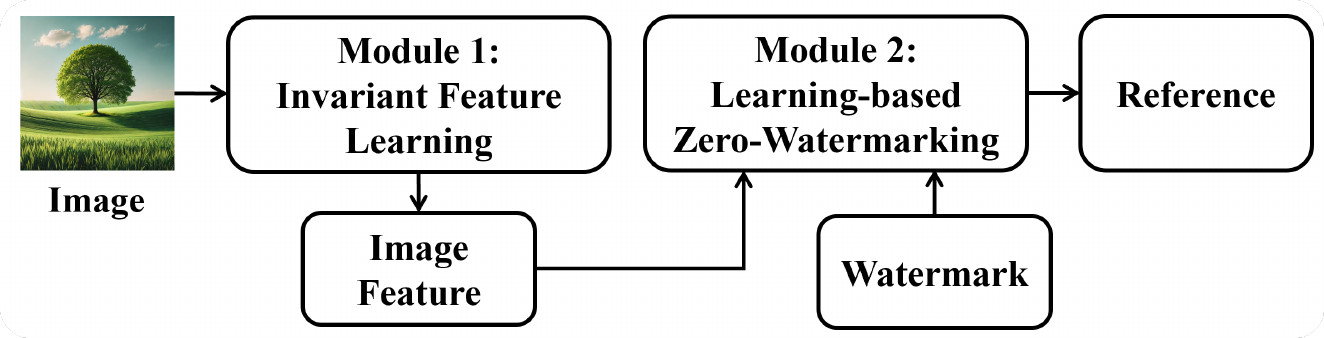}
    \vspace{-0.75em}
    \caption{Modules of the Proposed Method.}
    \label{fig:module}
    \vspace{-1.00em}
\end{figure}

To simplify the discussion, we conceptually decompose the proposed framework into two primary modules, as illustrated in Fig.~\ref{fig:module}. Module 1 performs invariant feature learning, where a deep network is trained to extract robust image features that remain stable under diverse distortions. Module 2 introduces a learning-based zero-watermarking scheme, which combines the extracted features with a given watermark to generate a reference signature for later extraction. The remainder of this section discusses the details of each module.

\subsection{Module 1: Invariant Feature Learning via Noise-Adversarial Training}
\begin{figure*}[!hb]
    \centering
    \vspace{-1.5em}
    \includegraphics[width=0.75\linewidth]{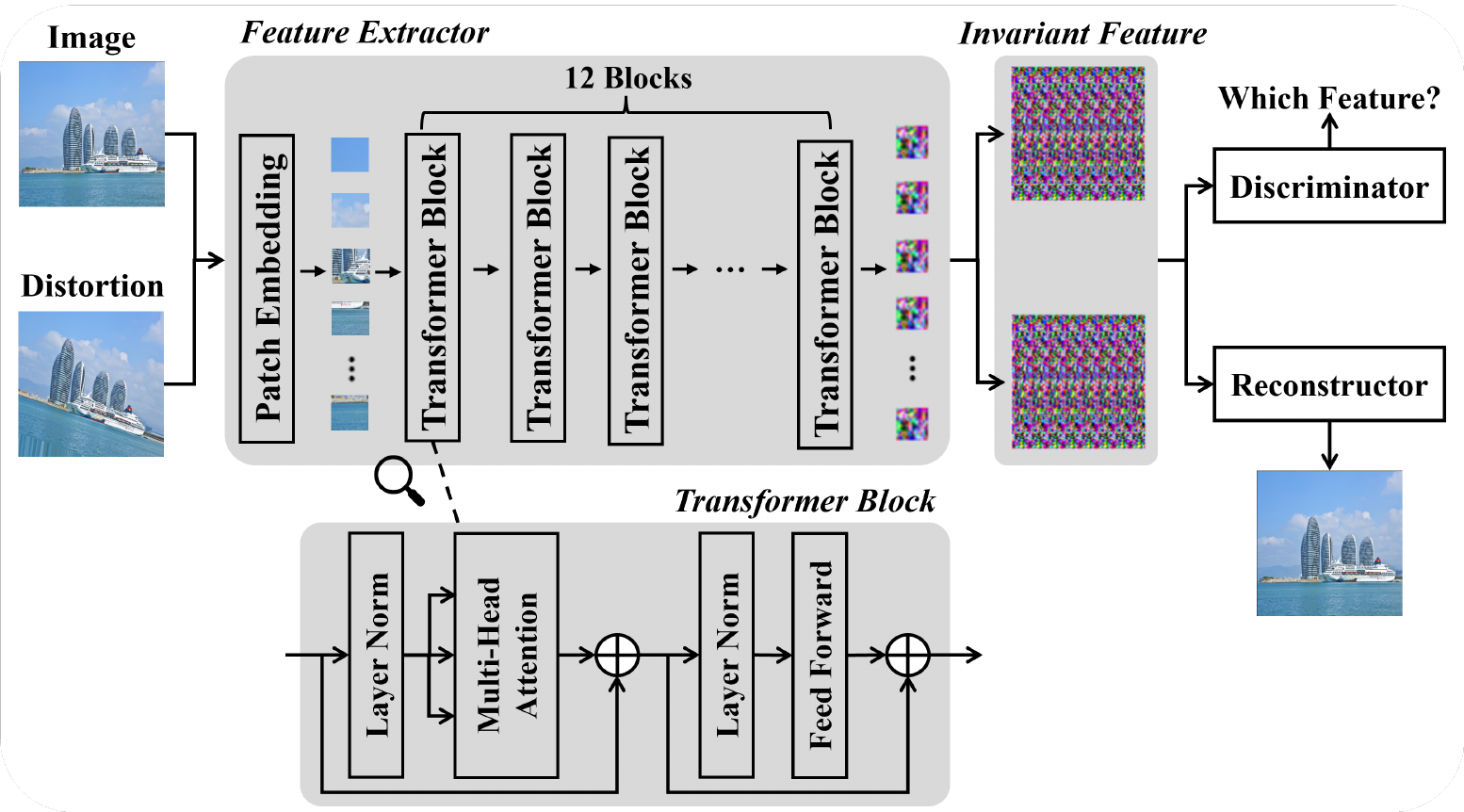}
    \vspace{-0.5em}
    \caption{Module 1: Proposed Invariant Feature Learning via Noise-Adversarial Training.}
    \label{fig:inv_feat}
\end{figure*}

Fig.~\ref{fig:inv_feat} illustrates the proposed invariant feature learning module based on noise-adversarial training. The framework consists of three main components: a feature extractor, a discriminator, and a reconstructor. Given both original and distorted images, the feature extractor, implemented using a Transformer-based architecture, generates high-level representations intended to be invariant to distortions. 
The discriminator is trained to classify whether a given feature originates from an original or distorted image, while the feature extractor is trained adversarially to fool the discriminator. 
This minimax interaction encourages the extractor to produce features that are indistinguishable across distortions, thereby promoting invariance. 
To ensure that the learned features retain semantic content, the reconstructor is trained to reconstruct the original image from these features, forcing the invariant features to preserve meaningful information relevant to the image itself.

\subsubsection{Feature Extractor}
The feature extractor \( \mathrm{FE} \) is implemented using the Vision Transformer (ViT) architecture~\cite{dosovitskiy2020image}, which is adapted here to learn features that are robust to image distortions. Given an input image \( x \in \mathbb{R}^{128 \times 128 \times 3} \), the image is divided into \( N = 64 \) non-overlapping patches, each of size \( 16 \times 16 \). Each patch \( p_i \) is flattened and projected into a \( d \)-dimensional latent space via a learnable linear transformation: \( z_i = W_p p_i + b_p \), where \( W_p \in \mathbb{R}^{d \times (16 \times 16 \times 3)} \) and \( b_p \in \mathbb{R}^d \). To encode positional information, learnable positional encodings \( E_{\text{pos}, i} \) are added to each patch embedding, resulting in \( z_i^{(0)} = z_i + E_{\text{pos}, i} \).

The sequence of embedded patches is then passed through \( L = 12 \) stacked Transformer blocks, each consisting of a multi-head self-attention (MHSA) layer followed by a feed-forward network (FFN). Formally, each block applies \( z^{(\ell)} = \mathrm{LayerNorm}(z^{(\ell-1)} + \mathrm{MHSA}(z^{(\ell-1)})) \), followed by \( z^{(\ell)} = \mathrm{LayerNorm}(z^{(\ell)} + \mathrm{FFN}(z^{(\ell)})) \), for \( \ell = 1, \dots, L \).


The output of the final Transformer block is a sequence of encoded patch tokens \( \{z_1^{(L)}, z_2^{(L)}, \dots, z_N^{(L)}\} \), which collectively represent the image in a latent feature space. Both the original and distorted images are passed through the same feature extractor, and the model is trained using adversarial objectives to ensure that these learned representations remain invariant to distortions.

To enable spatial supervision, the sequence of encoded patch tokens is further linearly projected and reshaped into a spatial feature tensor \( F_s \in \mathbb{R}^{128 \times 128 \times 3} \). This transformation is implemented via a linear projection applied to each patch token, followed by reshaping the sequence into a 2D grid aligned with the input resolution. The resulting spatial feature tensor is derived from the learned transformer representations (rather than the raw image) and is used as input to the discriminator and reconstructor modules described in the following subsections.

\subsubsection{Discriminator}

The discriminator is designed to classify whether a feature representation originates from an original or a distorted image. It follows a ViT-based architecture, structurally similar to the feature extractor, but adapted for binary classification.

Given the spatial feature tensor \(F_s \in \mathbb{R}^{128 \times 128 \times 3}\) derived from encoded patch features, the discriminator first divides it into 64 non-overlapping patches of size \(16 \times 16\). Each patch is flattened and linearly projected into a 64-dimensional latent space using a learnable embedding layer. To preserve spatial ordering, positional embeddings are added to each projected patch vector.

The sequence of patch embeddings is passed through a stack of \( L = 8 \) Transformer blocks. Each block contains a MHSA layer followed by a FFN as in Fig.~\ref{fig:inv_feat}, with residual connections and layer normalization applied at each submodule. This architecture allows the model to capture both local and global relationships across the image patches.

After Transformer processing, the output corresponding to the classification token is extracted and passed through a dense classification head with a softmax activation:
\(
D(F) = \mathrm{Softmax}(W_d \cdot F_{\text{CLS}} + b_d),
\)
where \( F_{\text{CLS}} \) denotes the final embedding of the classification token, and \( W_d \in \mathbb{R}^{1 \times d} \) and \( b_d \in \mathbb{R} \) are learnable parameters.

The discriminator is trained to distinguish between features derived from original and distorted images. Its gradients provide adversarial feedback to the feature extractor, encouraging the extractor to produce representations that are invariant to the applied distortions.

\subsubsection{Reconstructor}
\begin{figure*}[!ht]
    \centering
    \includegraphics[width=0.7\linewidth]{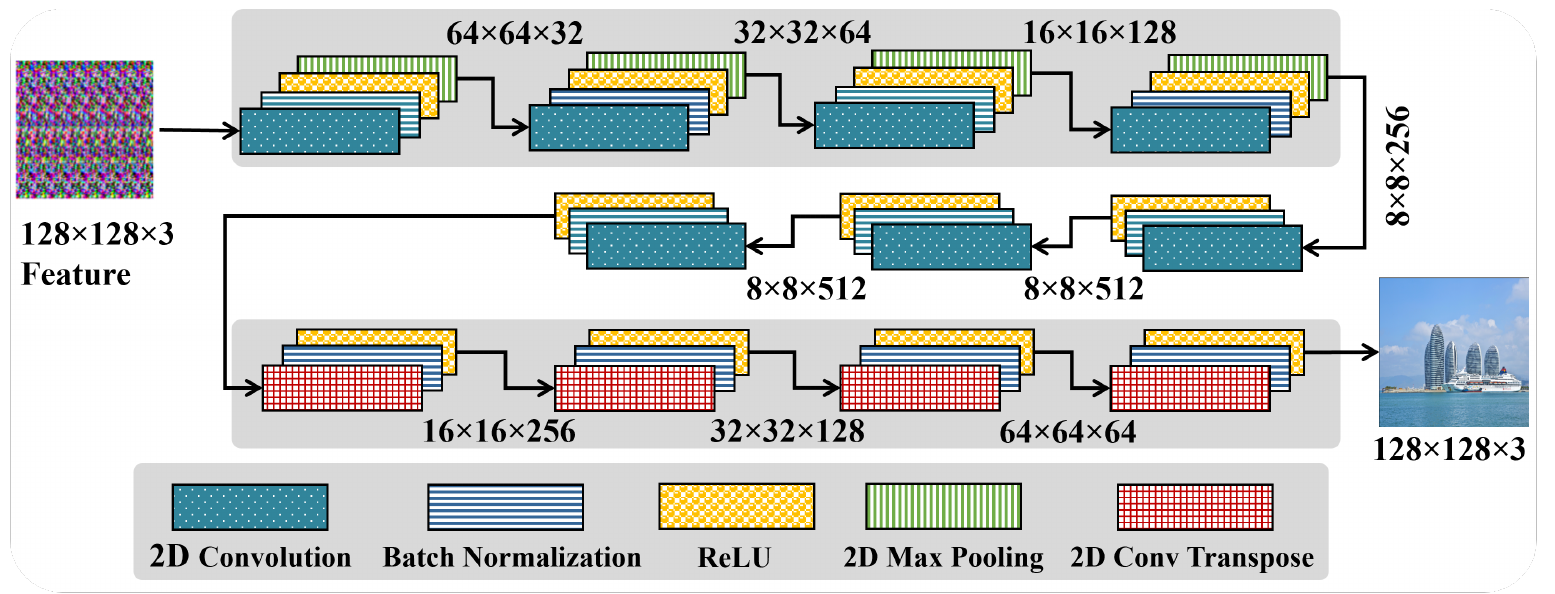}
    \vspace{-0.5em}
    \caption{Proposed Reconstructor Architecture.}
    \label{fig:recon}
    \vspace{-0.5em}
\end{figure*}

To prevent feature collapse, where the feature extractor outputs a constant or trivial representation (e.g., all zeros) regardless of the input, we introduce a reconstructor module \(R\) as an auxiliary constraint. While such representations may appear invariant, they are semantically meaningless and unusable for watermarking. By enforcing that the extracted features must support accurate reconstruction of the original image, we ensure that the representations retain semantic information in addition to distortion invariance. If the features collapse, reconstruction fails with high error, which the model is penalized for, thereby effectively discouraging degenerate solutions. 
The reconstructor, illustrated in Fig.~\ref{fig:recon}, can be conceptually described as an autoencoder-like architecture designed to enforce semantic preservation in the learned features. It consists of an encoder path, a bottleneck, and a decoder path. The input to the reconstructor is the feature tensor \( F \in \mathbb{R}^{128 \times 128 \times 3} \) produced by the feature extractor. 

The encoder comprises a series of convolutional blocks, each consisting of a 2D convolution layer followed by batch normalization and ReLU activation. These layers progressively reduce the spatial resolution from \( 128 \times 128 \) to \( 8 \times 8 \), while increasing the channel depth from 3 to 512. A 2D max pooling operation is applied at each stage to downsample the spatial dimensions. The resulting compressed representation at the bottleneck has dimensions \( 8 \times 8 \times 512 \), capturing high-level semantic information in a compact form. 
The decoder mirrors the encoder structure, utilizing transposed convolution layers to upsample the feature maps and gradually restore the spatial resolution. The channel dimensions are symmetrically reduced to reconstruct the image in its original shape. The final output of the decoder is a reconstructed image \( \hat{I} = f_{\text{recon}}(F) \), where a sigmoid activation is applied in the last layer to constrain pixel values between 0 and 1.

\subsubsection{Training}
The objective of the training process is to learn a feature extractor \( \mathrm{FE} \) that produces representations invariant to image distortions, while preserving semantic information necessary for downstream tasks such as watermarking. The training consists of three components: updating the discriminator \( D \), the feature extractor \( \mathrm{FE} \), and the reconstructor \( R \). Given an input image \( x \sim p(x) \), and its distorted version \( x' \sim p_T(x) \), we define the following loss functions and update routines.

\noindent\textbf{Training the Discriminator.}  
The discriminator \( D \) is trained to distinguish between features extracted from clean and distorted images. It is optimized using a binary cross-entropy loss:
\begin{align}
\mathcal{L}_D = & \; \mathbb{E}_{x \sim p(x)} \left[ \log D(\mathrm{FE}(x)) \right] \nonumber \\
& + \mathbb{E}_{x' \sim p_T(x)} \left[ \log \left( 1 - D(\mathrm{FE}(x')) \right) \right].
\end{align}
This objective encourages \( D \) to assign high confidence to features extracted from clean images and low confidence to those from distorted images.

\noindent\textbf{Training the Feature Extractor.} 
The feature extractor is trained adversarially to fool the discriminator, making the distributions of \( \mathrm{FE}(x) \) and \( \mathrm{FE}(x') \) indistinguishable. The adversarial loss is given by:
\begin{equation}
\mathcal{L}_{\mathrm{adv}} = \mathbb{E}_{x' \sim p_T(x)} \left[ \log \left( 1 - D(\mathrm{FE}(x')) \right) \right].
\end{equation}

To ensure that the features retain semantic information, the feature extractor is also optimized with a reconstruction loss via the reconstructor \( R \), which reconstructs the input from the features:
\(
\hat{x} = R(\mathrm{FE}(x)), \quad \hat{x}' = R(\mathrm{FE}(x')).
\)
The reconstruction loss is a weighted combination of the Mean Squared Error (MSE) and the Structural Similarity Index (SSIM):
\begin{equation}
\mathcal{L}_R = \lambda_S \left(1 - \mathrm{SSIM}(x, \hat{x})\right) + \lambda_M \| x - \hat{x} \|_2^2.
\label{eq: L_R}
\end{equation}
This loss is applied to both clean and distorted inputs, enforcing that features from either domain encode sufficient information for image recovery. 
The overall objective for the feature extractor combines the adversarial and reconstruction losses:
\begin{equation}
\mathcal{L}_{\mathrm{FE}} = \lambda_D \cdot \mathcal{L}_{\mathrm{adv}} + \mathcal{L}_R,
\end{equation}
where \( \lambda_D \) controls the strength of the adversarial term. By minimizing \( \mathcal{L}_{\mathrm{FE}} \), the feature extractor learns to produce invariant but semantically rich representations.

\noindent\textbf{Training the Reconstructor.}  
The reconstructor \( R \) is trained jointly with \( \mathrm{FE} \) to reconstruct the original image from the learned features. The same loss \( \mathcal{L}_R \) as in Eq.~(\ref{eq: L_R}) is minimized with respect to \( \theta_R \), the parameters of the reconstructor. 
This supervision ensures that the features do not collapse to trivial encodings and remain informative.
Algorithm~\ref{alg:training_inv} describes the proposed adversarial training for invariant feature learning. 

\begin{algorithm}[h]
\caption{Noise-Adversarial Training for Invariant Feature Learning}
\label{alg:training_inv}
\begin{algorithmic}[1]
\Require Feature Extractor \( \mathrm{FE} \), Discriminator \( D \), Reconstructor \( R \), transformation function \( T \), dataset \( X \), batch size \( B \), learning rate \( \alpha \), loss weights \( \lambda_1, \lambda_2 \)
\For{each epoch}
    \For{each batch of images \( x \) sampled from \( X \)}
        \State Generate distorted images: \( x' = T(x) \)
        \State Extract features: \( F = \mathrm{FE}(x) \), \( F' = \mathrm{FE}(x') \)

        \State \textbf{Train Discriminator:}
        \State Compute labels: \( y = 1 \) for \( F \), \( y = 0 \) for \( F' \)
        \State Compute discriminator loss \( \mathcal{L}_D \)
        \State Update \( D \) using \( \theta_D \leftarrow \theta_D - \alpha \nabla_{\theta_D} \mathcal{L}_D \)

        \State \textbf{Train Feature Extractor:}
        \State Compute adversarial loss \( \mathcal{L}_{\mathrm{adv}} \)
        \State Compute reconstruction loss \( \mathcal{L}_R \)
        \State Compute total loss \( \mathcal{L}_{\mathrm{FE}} = \lambda_1 \mathcal{L}_{\mathrm{adv}} + \lambda_2 \mathcal{L}_R \)
        \State Update \( \mathrm{FE} \) using \( \theta_{\mathrm{FE}} \leftarrow \theta_{\mathrm{FE}} - \alpha \nabla_{\theta_{\mathrm{FE}}} \mathcal{L}_{\mathrm{FE}} \)

        \State \textbf{Train Reconstructor:}
        \State Reconstruct images: \( \hat{x} = R(F) \)
        \State Compute reconstruction loss \( \mathcal{L}_R \)
        \State Update \( R \) using \( \theta_R \leftarrow \theta_R - \alpha \nabla_{\theta_R} \mathcal{L}_R \)
    \EndFor
    \State Save model weights every 10 epochs
\EndFor
\end{algorithmic}
\end{algorithm}

\subsubsection{Theoretical Perspective}
\label{sec:theory}
The proposed noise-adversarial training framework can be interpreted through two complementary perspectives: the Information Bottleneck (IB) principle~\cite{tishby2015deep} and domain adversarial learning~\cite{ganin2016domain}. 

From the perspective of the Information Bottleneck, the learning objective can be viewed as encouraging a representation \( F \) that retains information about the underlying image content while reducing sensitivity to distortion. This can be conceptually related to the trade-off:
\begin{equation}
\label{eqn: FE}
\max_{\mathrm{FE}} \, I(F, x) - \beta I(F, x'),
\end{equation}
where \( x \) denotes the original image and \( x' \) its distorted version. In our framework, this objective is not optimized explicitly. Instead, the adversarial loss encourages invariance by promoting similarity between feature distributions of \( x \) and \( x' \), while the reconstruction loss encourages the representation to preserve semantically relevant information.

From a domain adversarial learning perspective, the distortion process can be interpreted as inducing a domain shift. The original image \( x \) and its distorted version \( x' \) may be viewed as samples from related domains that share semantic structure but differ in nuisance factors. The adversarial objective encourages the feature extractor to produce representations that are difficult to distinguish across these domains. This can be interpreted as promoting approximate alignment between feature distributions:
\begin{equation}
\mathbb{E}_{x \sim p(x)} \left[ D(\mathrm{FE}(x)) \right] \approx 
\mathbb{E}_{x' \sim p_T(x)} \left[ D(\mathrm{FE}(x')) \right].
\end{equation}

However, enforcing invariance alone may lead to degenerate representations. The reconstruction constraint serves as a complementary mechanism that encourages the learned features to remain informative and structured:
\begin{equation}
x \approx R(\mathrm{FE}(x)), \quad x' \approx R(\mathrm{FE}(x')),
\end{equation}
thereby discouraging trivial or collapsed solutions. 
Taken together, these perspectives provide conceptual motivation for the proposed design, suggesting that the combination of adversarial alignment and reconstruction encourages the learning of invariant yet expressive feature representations for robust watermarking. 

\vspace{-0.75em}
\subsection{Module 2: Learning-based Zero-Watermarking with Invariant Feature}

In the second module, we present a novel multibit zero-watermarking scheme that leverages the invariant feature representation learned in Module 1. As a zero-watermarking scheme, our method does not modify the original image. Instead, a reference signature is optimized per image-watermark pair such that the semantic content captured by the invariant features can be reliably associated with a binary watermark message.

\subsubsection{Feature Extraction and Encoding}
Given an image \( x \), its invariant feature representation \( F = \mathrm{FE}(x) \in \mathbb{R}^{H \times W \times C} \) is extracted using the trained feature extractor from Module 1. To reduce the dimensionality and obtain a compact vector form, we apply a transformation function \( \Psi \), composed of global average pooling followed by a fully connected projection, yielding: 
\begin{equation}
\tilde{F} = \Psi(F), \quad \tilde{F} \in \mathbb{R}^{d}.
\end{equation}

\noindent Let \( W \in \{0,1\}^k \) be the target binary watermark message of length \( k \), and let \( C \in \mathbb{R}^{k \times d} \) denote a learnable reference signature matrix. The goal is to optimize \( C \) such that the inner product between each watermark bit vector \( C_i \in \mathbb{R}^d \) and the feature vector \( \tilde{F} \) aligns with the corresponding bit \( W_i \). The predicted watermark is computed as:
\begin{equation}
\hat{W}_i = \sigma(\tilde{F} \cdot C_i), \quad i = 1, \dots, k,
\end{equation}
where \( \sigma(\cdot) \) is the sigmoid activation function.

\subsubsection{Loss Function and Optimization}
To enforce alignment between the predicted watermark \( \hat{W} \) and the target message \( W \), we minimize a binary cross-entropy loss:
\begin{equation}
\mathcal{L}_W = - \sum_{i=1}^{k} \left[ W_i \log \hat{W}_i + (1 - W_i) \log (1 - \hat{W}_i) \right].
\end{equation}

\noindent To prevent overfitting and numerical instability, we regularize the reference signature matrix using an \( L_2 \) penalty:
\(
\mathcal{L}_C = \| C \|_2^2.
\)
The overall optimization objective becomes:
\(
\mathcal{L}_{\text{total}} = \mathcal{L}_W + \lambda_C \mathcal{L}_C,
\)
where \( \lambda_C \) controls the regularization strength.

The reference signature \( C \) is optimized using gradient descent:
\begin{equation}
C \leftarrow C - \eta \cdot \frac{\partial \mathcal{L}_{\text{total}}}{\partial C},
\end{equation}
where \( \eta \) is the learning rate. This optimization is performed per image-watermark pair during registration and does not involve updating the image or feature extractor. 
Similarly, the gradient of the total loss \( \mathcal{L}_{\text{total}} \) with respect to the parameters of the fully connected projection in \( \Psi \) is used to update the weights of this layer. This allows the transformation function \( \Psi \) to adapt to the watermark decoding task while keeping the feature extractor fixed.

\subsubsection{Watermark Extraction}
To extract the watermark, the distorted image is passed through the trained feature extractor and transformation function to yield \( \tilde{F}' = \Psi(\mathrm{FE}(x')) \). The predicted watermark is then computed as:
\begin{equation}
\hat{W}_i = \sigma(\tilde{F}' \cdot C_i), \quad i = 1, \dots, k.
\end{equation}

\noindent A thresholding step is applied to recover the binary message:
\begin{equation}
\tilde{W}_i = 
\begin{cases}
1, & \hat{W}_i \geq 0.5, \\
0, & \hat{W}_i < 0.5.
\end{cases}
\end{equation}

\noindent Robust watermark recovery is enabled by the invariance of \( \mathrm{FE} \), which ensures that \( \tilde{F}' \approx \tilde{F} \) even under substantial image distortions. Since the reference signature is optimized in feature space rather than image space, and the image is left unmodified, this approach fully conforms to the principles of zero-watermarking.

\begin{algorithm}[h]
\caption{Zero Watermarking and Extraction using Invariant Features}
\label{alg:zw_inv}
\begin{algorithmic}[1]
\Require Original images $\{I_i\}_{i=1}^{N}$, watermarks $\{W_i\}_{i=1}^{N}$, reference tensor $C \in \mathbb{R}^{N \times B \times d}$, feature extractor $\mathrm{FE}$
\Ensure Optimized reference tensor $C$ and projection module $\Psi$

\For{each image $I_i$}
    \State Extract invariant feature map: $F_i = \mathrm{FE}(I_i)$
    \State Compute projected feature: $\tilde{F}_i = \Psi(F_i),\ \tilde{F}_i \in \mathbb{R}^d$
    \For{$b = 1$ to $B$}
        \State Compute dot product: $s_{i,b} = \tilde{F}_i^\top C_{i,b}$
    \EndFor
    \State Apply sigmoid: $\hat{W}_i = \sigma(s_i) \in \mathbb{R}^B$
    \State Compute BCE loss: $\mathcal{L}_W = \mathrm{BCE}(\hat{W}_i, W_i)$
    \State Regularize reference: $\mathcal{L}_C = \|C\|_2^2$
    \State Total loss: $\mathcal{L}_{\text{total}} = \mathcal{L}_W + \mathcal{L}_C$
    \State Update $C$: $\theta_C \leftarrow \theta_C - \alpha \nabla_{\theta_C}\mathcal{L}_{\text{total}}$
    \State Update $\Psi$: $\theta_\Psi \leftarrow \theta_\Psi - \alpha \nabla_{\theta_\Psi}\mathcal{L}_{\text{total}}$
\EndFor

\Statex
\State \textbf{Watermark Extraction}
\State \textbf{Input:} Noisy image $I'$, $\mathrm{FE}$, optimized $C$, optimized $\Psi$
\State Extract feature: $F' = \mathrm{FE}(I')$
\State Project: $\tilde{F}' = \Psi(F')$
\For{$b = 1$ to $B$}
    \State Compute: $s'_b = (\tilde{F}')^\top C_b$
    \State Apply sigmoid: $\hat{W}_b = \sigma(s'_b)$
    \If{$\hat{W}_b \ge 0.5$}
        \State $W_b \gets 1$
    \Else
        \State $W_b \gets 0$
    \EndIf
\EndFor
\State \Return $W$
\end{algorithmic}
\end{algorithm}

\subsubsection{Method Advantages}
A key practical advantage of the proposed framework is its flexibility in associating multiple watermark messages with a single image. Since the reference signature C is optimized in the invariant feature space rather than embedded in the image, it can be independently learned for each pair of image-watermarks without modifying the original image or retraining the feature extractor. In our formulation, a reference signature tensor of size $128 \times 128 \times 3$ is stored for each such pair. 
This enables a natural one-to-many mapping, where a single image can be linked to multiple watermark identities, each defined by a distinct reference signature. The storage requirement scales linearly with the number of assigned watermarks. During verification, the feature extracted from a query image is compared against candidate reference signatures, and only the correct signature produces a successful match, allowing reliable watermark identification and supporting user-specific ownership or multi-level authorization.

From a theoretical perspective, the proposed approach can be interpreted as solving a constrained optimization problem in the invariant feature space. Given a fixed, distortion-invariant representation \( \tilde{F} \), the training of \( C \) seeks to project \( \tilde{F} \) onto a set of learned directional codes \( \{C_1, \dots, C_k\} \) such that their inner products reflect a target binary vector \( W \). The sigmoid activation and binary cross-entropy loss induce a soft margin between watermark bits, effectively identifying a hyperplane that separates \( \tilde{F} \cdot C_i \geq 0.5 \) from \( < 0.5 \). By minimizing this loss, the solution converges to a directionally optimal configuration of \( C \), tailored to the unique geometry of the input feature. This learning-based formulation generalizes well to unseen transformations, as the optimization finds reference codes that are not just memorized, but robustly aligned with the semantic content captured by \( \tilde{F} \).

%% file: sec/4_experimental_result.tex
\vspace{-0.25em}
\section{Experimental Result}
\label{sec:exp_result}

This section presents evaluation of the proposed framework. Section~\ref{sec:data} describes the datasets and preprocessing and augmentation strategies used for training and evaluation. 
Section~\ref{sec:inv_analyasis} analyzes the quality and stability of the learned invariant features, including feature consistency under distortion and the quality of image reconstruction. 
Section~\ref{sec:inv_robust_compare} compares our method with existing self-supervised feature extractors, such as SimCLR, BYOL, and VICReg, using cosine similarity and linear evaluation metrics to assess feature robustness.
Finally, Section~\ref{sec:wm} demonstrates the effectiveness of our learning-based zero-watermarking scheme, providing comparisons with competitive baselines under a range of distortions.

\vspace{-0.5em}
\subsection{Dataset and Feature Training}
\label{sec:data}
To evaluate the effectiveness and robustness of the proposed framework, we conducted experiments across multiple datasets, each selected to serve a specific purpose. All images were resized to a resolution of \(128 \times 128 \times 3\) and normalized to the range \([0, 1]\) to ensure consistency and stable convergence during training.

For invariant feature learning and analysis (Section~\ref{sec:inv_analyasis}), we used a subset of 40,000 images from the MSCOCO dataset, which contains a diverse range of real-world scenes. To assess generalization, we evaluated the trained models on 5,000 MSCOCO images, as well as 5,000 images each from the Imagenette and Flickr30k datasets. These datasets introduce unseen variations in content and background, helping to validate the stability of the learned representations under domain shifts. Additionally, high-resolution robustness was tested using 900 images from the DIV2K dataset.

For linear evaluation (Section~\ref{sec:inv_robust_compare}), we used 10,000 labeled images from the CIFAR-100 dataset to train a downstream classifier on frozen features, and evaluated it on a separate test set of 3,000 images. CIFAR-100 provides a well-established benchmark for assessing semantic discriminability of features across object categories.

\begin{figure}[!ht]
    \centering
    \vspace{-1.0em}
    \includegraphics[width=0.7\linewidth]{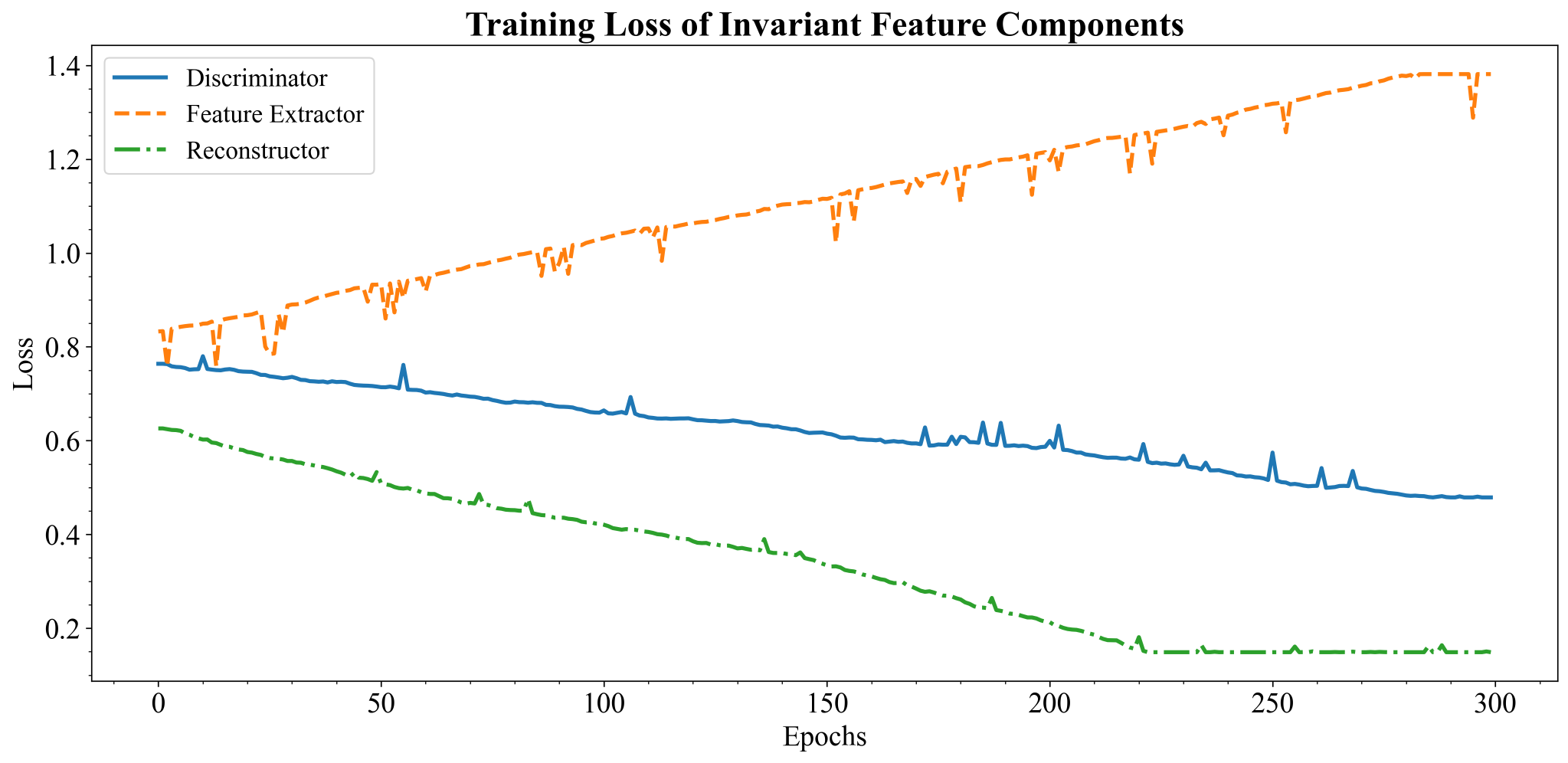}
    \vspace{-1.0em}
    \caption{Training Loss Curves for Invariant Feature Learning.}
    \label{fig:inv_training_loss}
    \vspace{-1.0em}
\end{figure}

The feature extractor was trained using the Adam optimizer with separate learning rates for each component: \( \eta_{\mathrm{FE}} = 0.0005 \) for the feature extractor, \( \eta_D = 0.001 \) for the discriminator, and \( \eta_R = 0.0001 \) for the reconstructor. 
As shown in Figure~\ref{fig:inv_training_loss}, the training loss curves indicate stable and consistent convergence across all components of our model. The Discriminator and Reconstructor losses decrease progressively, reflecting improved capability in distinguishing invariant features and reconstructing original content. Meanwhile, the Feature Extractor shows a steady increase in performance, suggesting it is learning more robust and invariant representations over time. The smooth convergence of these loss curves highlights the effectiveness of our training strategy and suggests that the model is becoming more accurate and generalizable.

Since the proposed framework learns and generalizes invariance through input distortions, we apply a range of geometric and photometric transformations during training and testing. Table~\ref{tab:augmentations} summarizes the distortion strategies used. Notably, during training, only geometric distortions (e.g., rotation, translation, shear) are applied, while at test time, a broader spectrum of distortions, including blurring, compression, and noise, is introduced. To ensure that image rotation does not result in content loss, we apply a combined rotation and resizing operation that preserves the full image content within the frame.This design enables us to evaluate the generalization capacity of the learned invariant features: although trained only with geometric distortions, the model demonstrates invariance to a wide variety of unseen image processing attacks such as salt-and-pepper noise and JPEG compression.

\vspace{-0.5em}
\begin{table}[h]
    \centering
    \caption{Summary of Distortions in Training and Testing.}
    \label{tab:augmentations}
    \vspace{0.5em}
    \resizebox{0.4\linewidth}{!}{
    \begin{tabular}{|l|c|c|}
        \hline
        \textbf{Distortion Type} & \textbf{Training} & \textbf{Testing} \\
        \hline
        Rotation & $\pm$15° & $\pm$25° \\
        Width Shift & 15\% & 25\% \\
        Height Shift & 15\% & 25\% \\
        Shear & 15\% & 25\% \\
        Zoom & 15\% & 25\% \\
        Horizontal Flip & Yes & Yes \\
        Gaussian Blur & No & Yes \\
        Solarization & No & Yes \\
        Crop / Cutout & No & Yes \\
        JPEG Compression & No & Yes \\
        Brightness Adjustment & No & Yes \\
        Contrast Adjustment & No & Yes \\
        Hue / Saturation Adjustment & No & Yes \\
        Gaussian Noise & No & Yes \\
        Salt \& Pepper Noise & No & Yes \\
        \hline
    \end{tabular}
    }
\end{table}

\subsection{Analysis of the Proposed Invariant Feature}
\label{sec:inv_analyasis}
To assess the effectiveness of the proposed invariant feature learning framework, we conduct a detailed analysis of feature robustness and reconstruction quality across a variety of datasets and distortions. We divide the distortions into two categories: (1) common image processing distortions, such as blur, noise, and compression, and (2) geometric transformations, such as rotation, translation, and zoom. Our evaluation is performed on four datasets: MSCOCO, ImageNet, Flickr30k, and DIV2K, selected to cover a wide spectrum of image content and resolutions.

To quantify feature invariance, we compute the cosine similarity between the invariant feature vectors extracted from the original image \( x \) and a distorted version \( x' \). A higher similarity indicates stronger invariance to the applied transformation. 
Quality of the the proposed reconstruction is evaluated using the peak signal-to-noise ratio (PSNR) between the original image \( x \) and its reconstruction \( \hat{x} \) from the extracted features. These two metrics together offer a comprehensive view of both the stability and informativeness of the learned representations.

Table~\ref{tab:cosine_psnr_image_attacks} reports the cosine similarity and PSNR under common image processing distortions. Despite being trained only with geometric augmentations, the model demonstrates strong generalization to unseen distortions. For instance, cosine similarity remains consistently high under blur and JPEG compression ($\geq$0.93 across all datasets), indicating minimal impact on the feature representation. Gaussian noise and solarization result in modest drops in similarity, while salt-and-pepper noise introduces the largest variation, particularly on MSCOCO (0.92). Notably, Flickr30k and ImageNet show higher resilience to these pixel-level attacks, with cosine similarity values reaching up to 0.96.

The PSNR scores corroborate the feature analysis. MSCOCO, ImageNet, and Flickr30k achieve high reconstruction fidelity under most distortions, with blur and JPEG yielding PSNR values above 30 dB. In contrast, DIV2K, due to its higher resolution and finer detail, shows more degradation under extreme perturbations, particularly salt-and-pepper noise (27.3 dB) and solarization (26.0 dB). These results confirm that the learned features are both distortion-invariant and semantically expressive across multiple content domains.


\begin{table*}[!hbt] 
\caption{Feature cosine similarity and image reconstruction PSNR under common distortions (mean $\pm$ std over 5 runs).}
\vspace{0.5em}
\label{tab:cosine_psnr_image_attacks} 
\resizebox{0.98\linewidth}{!}{
\begin{tabular}{|l|ccccc|ccccc|}
\hline
\textbf{Dataset} & \multicolumn{5}{c|}{\textbf{Cosine Similarity $\uparrow$}} & \multicolumn{5}{c|}{\textbf{PSNR (dB) $\uparrow$}} \\ 
\cline{2-11}
& \textbf{Blur} & \textbf{JPEG} & \textbf{Gaussian Noise} & \textbf{Salt \& Pepper} & \textbf{Solarization} & \textbf{Blur (1.5)} & \textbf{JPEG} & \textbf{Gaussian Noise} & \textbf{Salt \& Pepper} & \textbf{Solarization} \\ 
\hline
MSCOCO & $0.97 \pm 0.014$ & $0.96 \pm 0.024$ & $0.96 \pm 0.032$ & $0.92 \pm 0.037$ & $0.93 \pm 0.021$ 
       & $32.5 \pm 0.41$ & $31.2 \pm 0.68$ & $30.9 \pm 0.83$ & $28.8 \pm 1.12$ & $29.5 \pm 0.52$ \\ 

ImageNet & $0.95 \pm 0.031$ & $0.94 \pm 0.038$ & $0.93 \pm 0.041$ & $0.96 \pm 0.019$ & $0.92 \pm 0.036$ 
         & $33.9 \pm 0.57$ & $30.8 \pm 1.21$ & $31.5 \pm 0.92$ & $29.2 \pm 1.48$ & $30.3 \pm 1.05$ \\ 

Flickr30k & $0.98 \pm 0.011$ & $0.96 \pm 0.026$ & $0.95 \pm 0.043$ & $0.96 \pm 0.031$ & $0.95 \pm 0.022$ 
          & $32.2 \pm 0.93$ & $31.8 \pm 0.74$ & $29.8 \pm 1.96$ & $26.5 \pm 1.78$ & $30.2 \pm 0.88$ \\ 

DIV2K & $0.95 \pm 0.028$ & $0.93 \pm 0.036$ & $0.93 \pm 0.039$ & $0.91 \pm 0.044$ & $0.93 \pm 0.033$ 
      & $29.5 \pm 1.12$ & $31.0 \pm 0.91$ & $28.7 \pm 1.47$ & $27.3 \pm 1.18$ & $26.0 \pm 1.32$ \\  
\hline
\end{tabular}
}
\end{table*}

Table~\ref{tab:cosine_psnr_geometric_attacks} presents the results under geometric transformations. Rotation shows the highest feature invariance across all datasets (cosine similarity $\geq$0.95), followed by width and height shifts. Shear and zoom introduce slightly more disruption, with cosine similarity values dropping to 0.91–0.94 for ImageNet and DIV2K. Among geometric distortions, zoom exhibits the most pronounced impact on feature consistency, likely due to scale-dependent feature extraction challenges.

Reconstruction quality remains high for most geometric transformations, with PSNR values frequently exceeding 30 dB. For example, width shift on ImageNet achieves 33.5 dB, while even under rotation, MSCOCO retains a PSNR of 33.5 dB. However, zoom transformations reduce PSNR across all datasets, especially for Flickr30k (28.6 dB), indicating that scale variations affect both the spatial detail and feature structure more than other affine transformations.


\begin{table*}[!hbt] 
\caption{Feature cosine similarity and image reconstruction PSNR under geometric transformations (mean $\pm$ std over 5 runs).} 
\vspace{0.5em}
\label{tab:cosine_psnr_geometric_attacks} 
\resizebox{0.98\linewidth}{!}{
\begin{tabular}{|l|ccccc|ccccc|}
\hline
\textbf{Dataset} & \multicolumn{5}{c|}{\textbf{Cosine Similarity $\uparrow$}} & \multicolumn{5}{c|}{\textbf{PSNR (dB) $\uparrow$}} \\ 
\cline{2-11}
& \textbf{Rotation} & \textbf{Width Shift} & \textbf{Height Shift} & \textbf{Shear} & \textbf{Zoom} 
& \textbf{Rotation} & \textbf{Width Shift} & \textbf{Height Shift} & \textbf{Shear} & \textbf{Zoom} \\ 
\hline
MSCOCO & $0.97 \pm 0.012$ & $0.96 \pm 0.018$ & $0.96 \pm 0.017$ & $0.94 \pm 0.021$ & $0.95 \pm 0.019$ 
       & $33.5 \pm 0.38$ & $31.1 \pm 0.56$ & $31.6 \pm 0.49$ & $32.3 \pm 0.61$ & $29.6 \pm 0.74$ \\ 

ImageNet & $0.95 \pm 0.019$ & $0.94 \pm 0.024$ & $0.94 \pm 0.027$ & $0.91 \pm 0.033$ & $0.93 \pm 0.026$ 
         & $31.9 \pm 0.64$ & $33.5 \pm 0.51$ & $29.2 \pm 0.97$ & $31.8 \pm 0.68$ & $29.1 \pm 0.86$ \\ 

Flickr30k & $0.97 \pm 0.013$ & $0.95 \pm 0.019$ & $0.94 \pm 0.028$ & $0.94 \pm 0.022$ & $0.92 \pm 0.027$ 
          & $30.2 \pm 0.94$ & $33.7 \pm 0.63$ & $31.7 \pm 1.21$ & $30.2 \pm 0.88$ & $28.6 \pm 1.09$ \\ 

DIV2K & $0.96 \pm 0.021$ & $0.92 \pm 0.030$ & $0.93 \pm 0.027$ & $0.92 \pm 0.031$ & $0.94 \pm 0.024$ 
      & $31.5 \pm 0.82$ & $32.1 \pm 0.71$ & $30.8 \pm 1.18$ & $32.8 \pm 0.93$ & $29.6 \pm 1.11$ \\ 
\hline
\end{tabular}
}
\end{table*}

\begin{figure}[ht]
    \centering
    \vspace{-0.75em}
    \includegraphics[width=0.4\linewidth]{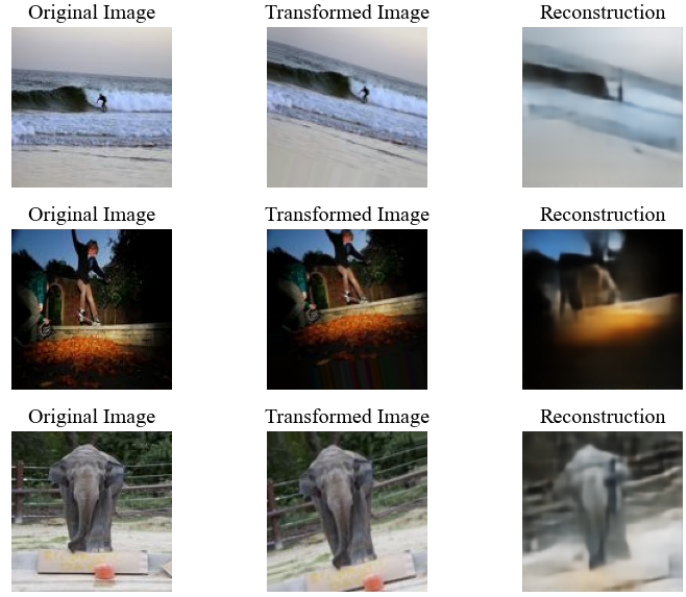} 
    \vspace{-1.00em}
    \caption{Representative Examples of Image Reconstruction From Invariant Features.}
    \label{fig:example}
    \vspace{-1.00em}
\end{figure}

The results show that the proposed framework effectively learns invariant features that generalize well to unseen distortions, including common image processing and geometric transformations. This robustness stems from the adversarial training setup, where the generator produces distortion-invariant features that the discriminator cannot distinguish. By aligning features across original and distorted images, the model learns to capture semantic content rather than distortion-specific cues. As a result, the features remain stable even under novel distortions at test time, enabling reliable watermark verification. To demonstrate that these features retain semantic information, reconstruction results are shown in Figure~\ref{fig:example}.

\subsection{Comparative Analysis of the Proposed Invariant Feature}
\label{sec:inv_robust_compare}
To evaluate the robustness and effectiveness of the proposed invariant feature learning framework, we compare it with five representative self-supervised learning (SSL) baselines: SimCLR~\cite{chen2020simple}, BYOL~\cite{grill2020bootstrap}, Barlow Twins~\cite{zbontar2021barlow}, VICReg~\cite{bardes2021vicreg}, and VIBCREG~\cite{lee2021vibcreg}. These methods were selected for their strong performance across SSL benchmarks and for covering diverse learning paradigms, including contrastive learning, redundancy reduction, and variance-invariance-covariance regularization. 
We assess the learned representations along two key axes: (1) feature invariance, i.e., the stability of feature embeddings under distortions, and (2) semantic descriptiveness, i.e., how well the features support classification when linearly probed. These two criteria together offer a comprehensive measure of robustness and utility for downstream watermarking tasks.

\subsubsection{Invariant Feature Similarity}
We first evaluate the degree to which each method preserves feature consistency under perturbations. Given an input image and a distorted variant, we extract feature embeddings using each method and compute the cosine similarity between the original and distorted features. A higher similarity indicates stronger invariance to the applied transformation. We test against five common distortions: Gaussian blur, JPEG compression, Gaussian noise, salt \& pepper noise, and hue shift.


Although self-supervised learning (SSL) models provide strong general-purpose visual representations, their invariance is primarily driven by predefined augmentations and does not explicitly target distortion-specific variability that affects watermark consistency. As a result, SSL features may still exhibit instability under distortions such as compression, filtering, and geometric transformations. 

In contrast, the proposed framework explicitly enforces distortion invariance through adversarial training, which minimizes feature discrepancies between clean and distorted inputs. At the same time, the reconstruction constraint ensures that the learned features retain sufficient semantic information and avoid collapse. This targeted invariance learning leads to more stable and reliable feature representations for multi-bit zero-watermark decoding under diverse distortions.

Table~\ref{tab:cosine_similarity_ssl} reports the results across four datasets. The proposed method consistently outperforms the SSL baselines across all distortions and datasets. For instance, under Gaussian blur and JPEG compression, our model achieves cosine similarities up to 0.98, demonstrating high resistance to low-level degradations. Even under challenging conditions like salt \& pepper noise and hue shifts, our model maintains higher feature consistency than all compared methods. These results validate the effectiveness of our adversarial invariant learning objective in producing distortion-invariant feature embeddings.

\begin{table*}[!htb] 
\centering 
\caption{Quantitative Comparison of Feature Invariance (Cosine Similarity) across SSL Methods Under Diverse Distortions on Multiple Datasets.}
\vspace{0.5em}
\label{tab:cosine_similarity_ssl} 
\resizebox{0.90\linewidth}{!}{%
\begin{tabular}{|l|c|ccccc|}
\hline
\multirow{2}{*}{\textbf{Dataset}} & \multirow{2}{*}{\textbf{Method}} & \multicolumn{5}{c|}{\textbf{Cosine Similarity (Feature Invariance) $\uparrow$}} \\ 
\cline{3-7}
& & Gaussian Blur (1.5) & JPEG (15) & Gaussian Noise (0.15) & Salt \& Pepper (0.15) & Hue (0.2) \\ 
\hline
\multirow{6}{*}{MSCOCO} 
& SimCLR & 0.97 & 0.92 & 0.92 & \textbf{0.94} & 0.90 \\ 
& VICReg & 0.92 & 0.86 & 0.84 & 0.91 & 0.89 \\ 
& Barlow Twins & 0.93 & 0.88 & 0.87 & 0.90 & 0.92 \\ 
& BYOL & 0.96 & 0.93 & 0.91 & 0.93 & \textbf{0.95} \\ 
& VIbCReg & 0.94 & 0.95 & 0.91 & 0.88 & 0.91 \\ 
& Ours & \textbf{0.98} & \textbf{0.96} & \textbf{0.95} & 0.93 & 0.93 \\ 
\hline

\multirow{6}{*}{ImageNet} 
& SimCLR & 0.94 & 0.87 & 0.89 & 0.92 & 0.90 \\ 
& VICReg & 0.92 & 0.89 & 0.92 & 0.93 & \textbf{0.92} \\ 
& Barlow Twins & 0.93 & 0.90 & \textbf{0.96} & 0.91 & 0.91 \\ 
& BYOL & 0.94 & 0.91 & 0.92 & \textbf{0.95} & 0.89 \\ 
& VIbCReg & 0.95 & 0.92 & 0.88 & 0.93 & 0.90 \\ 
& Ours & \textbf{0.96} & \textbf{0.94} & 0.94 & \textbf{0.95} & \textbf{0.92} \\ 
\hline

\multirow{6}{*}{Flickr30k} 
& SimCLR & 0.97 & \textbf{0.96} & 0.93 & 0.93 & 0.94 \\ 
& VICReg & 0.95 & 0.93 & \textbf{0.95} & 0.92 & 0.92 \\ 
& Barlow Twins & 0.94 & 0.93 & 0.94 & 0.94 & 0.93 \\ 
& BYOL & 0.95 & 0.92 & 0.98 & 0.93 & \textbf{0.95} \\ 
& VIbCReg & 0.96 & 0.93 & 0.94 & 0.94 & 0.91 \\ 
& Ours & \textbf{0.98} & \textbf{0.96} & 0.94 & \textbf{0.95} & \textbf{0.95} \\ 
\hline

\multirow{6}{*}{DIV2K} 
& SimCLR & \textbf{0.95} & 0.88 & 0.91 & 0.87 & 0.91 \\ 
& VICReg & 0.94 & 0.92 & 0.90 & 0.93 & 0.92 \\ 
& Barlow Twins & 0.94 & 0.91 & 0.92 & 0.87 & 0.92 \\ 
& BYOL & \textbf{0.95} & \textbf{0.93} & 0.90 & 0.90 & 0.91 \\ 
& VIbCReg & 0.93 & 0.90 & 0.91 & 0.91 & 0.90 \\ 
& Ours & \textbf{0.95} & \textbf{0.93} & \textbf{0.94} & \textbf{0.92} & \textbf{0.93} \\ 
\hline
\end{tabular}%
}
\end{table*}

\subsubsection{Linear Evaluation for Semantic Discriminability}
To assess whether the learned features also retain semantic richness, we conduct a linear evaluation protocol. For each SSL method, including ours, we extract features from a frozen encoder and train a linear classifier on 10,000 clean CIFAR-100 training images. The classifiers are then evaluated on distorted test sets from both CIFAR-100 and Imagenette, using the same five distortion types.

\begin{figure*}[!htb]
    \centering
    \vspace{-0.50em}
    \includegraphics[width=0.375\linewidth]{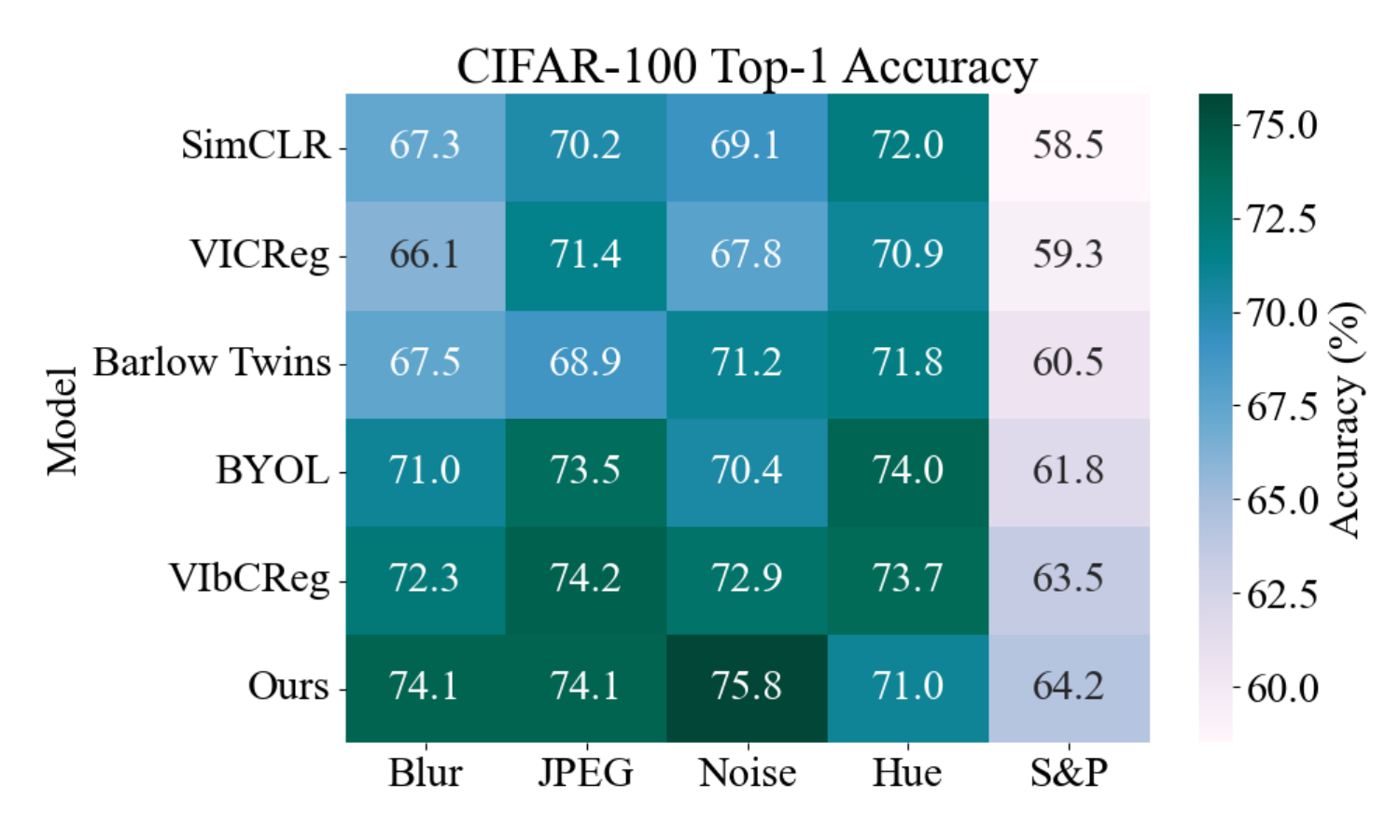}
    \includegraphics[width=0.375\linewidth]{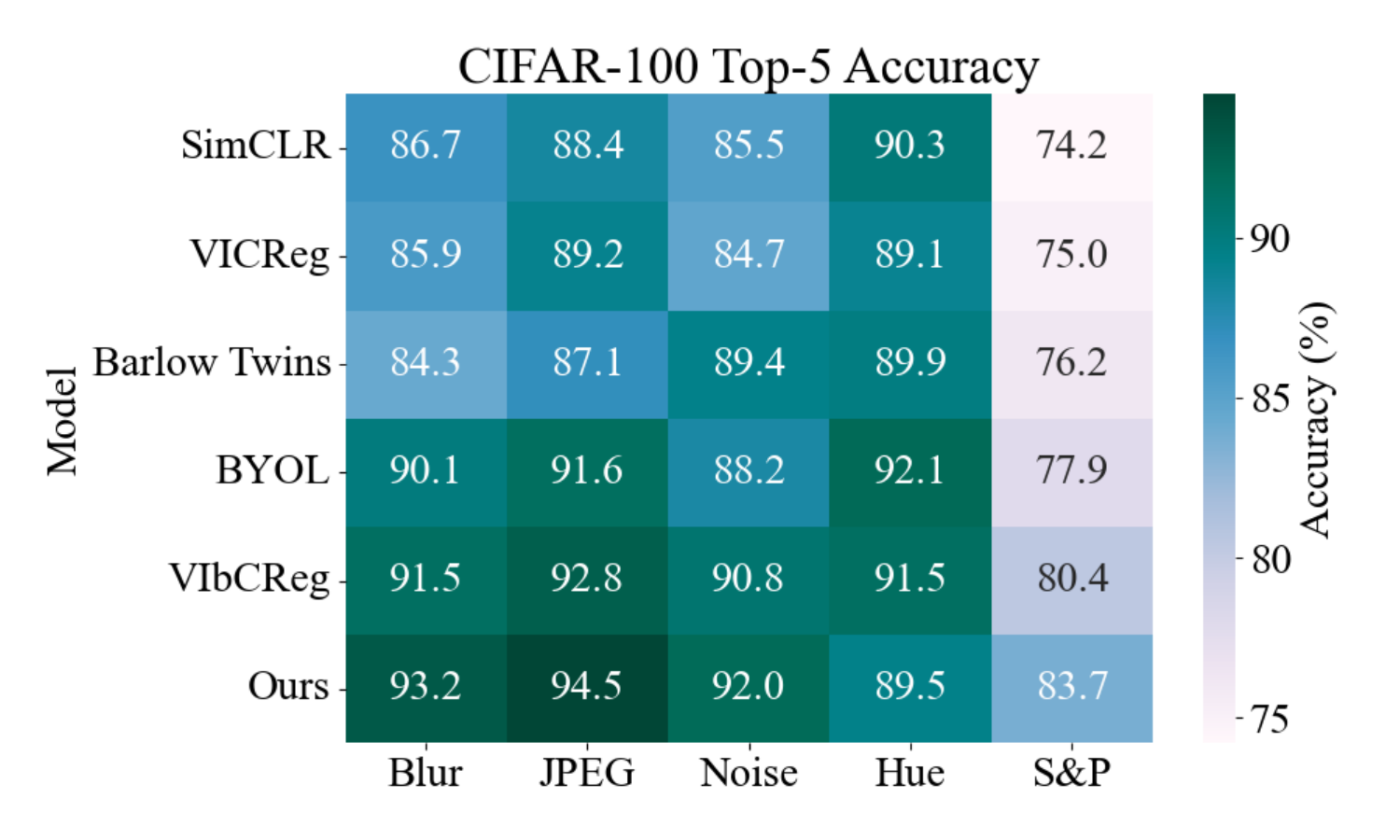}
    \includegraphics[width=0.375\linewidth]{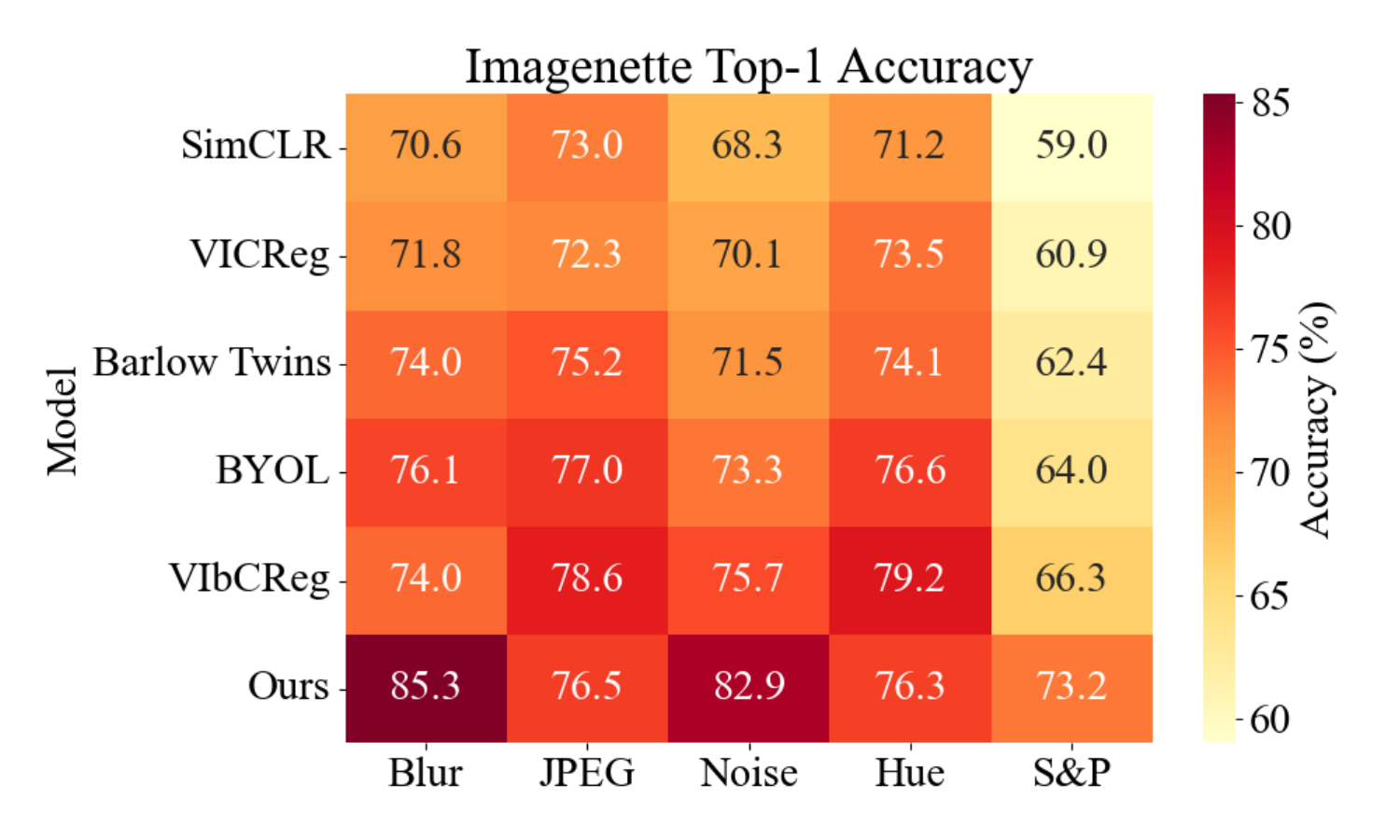}
    \includegraphics[width=0.375\linewidth]{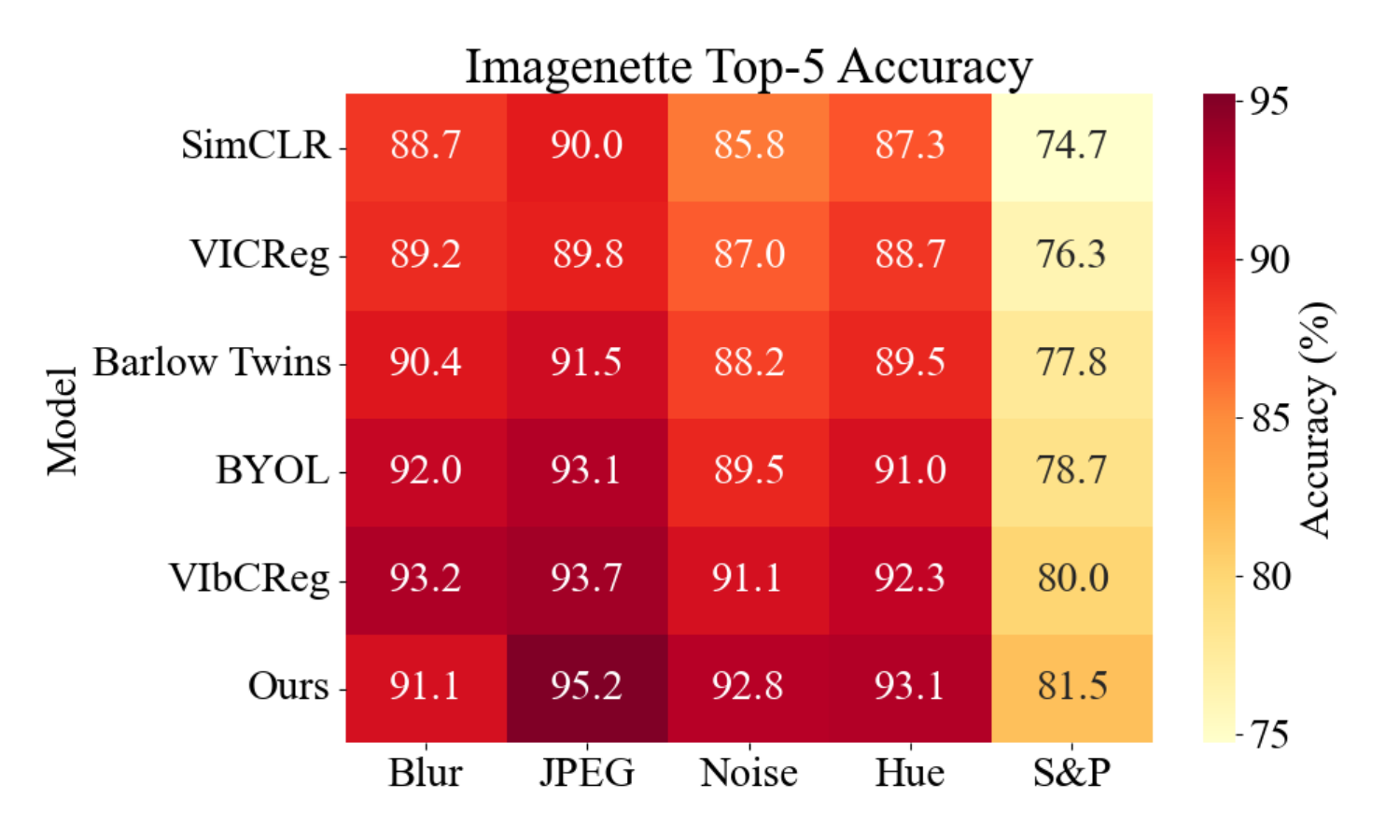}
    \vspace{-0.50em}
    \caption{Linear Evaluation of Representation Quality: Top-1 and Top-5 Classification Accuracy Across SSL Methods on CIFAR-100 and Imagenette Under Various Distortions.}
    \label{fig:heatmaps_ssl_accuracy}
    \vspace{-0.75em}
\end{figure*}

\begin{figure}[ht]
\centering
\vspace{-0.75em}
\includegraphics[width=0.75\linewidth]{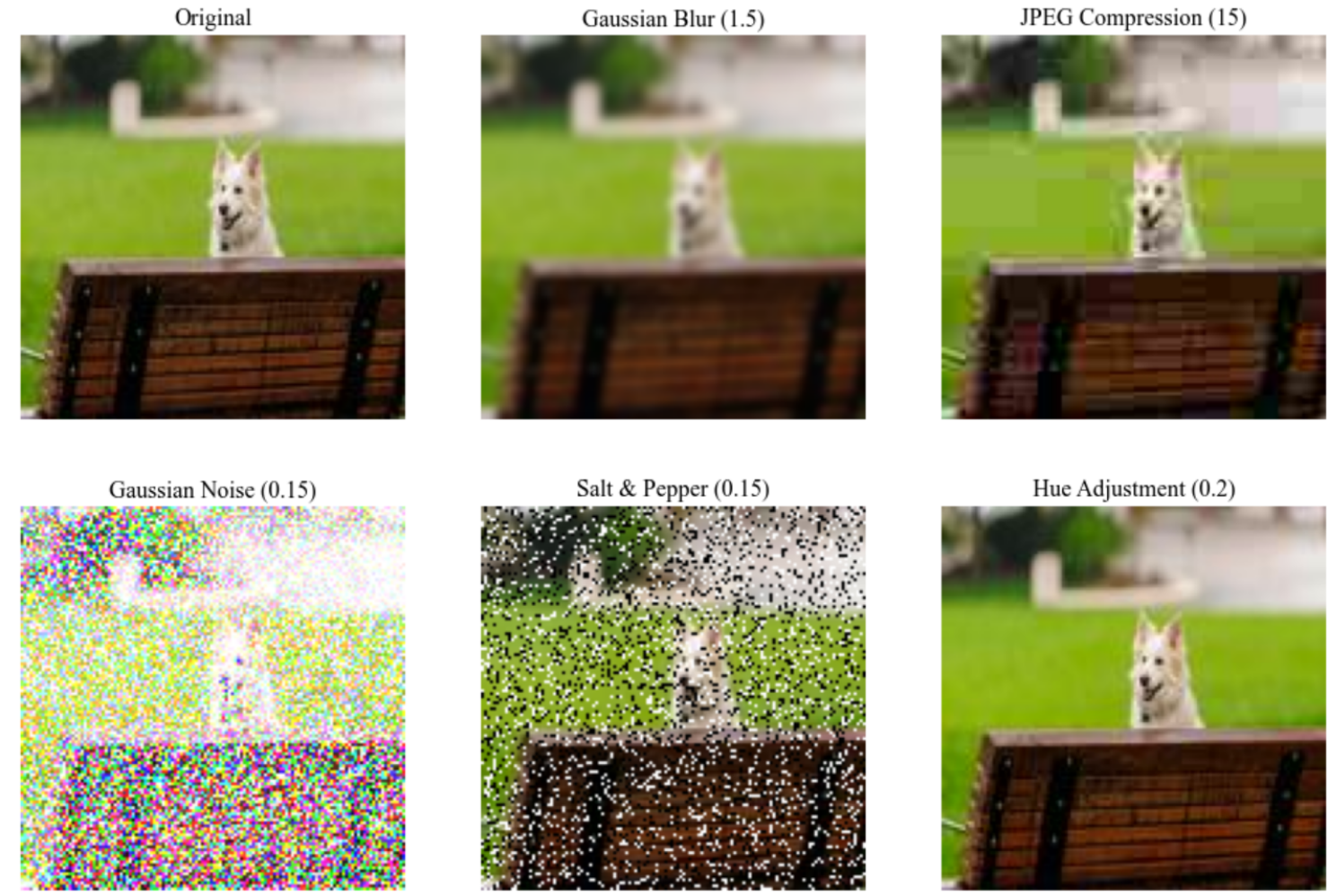}
\caption{Visual Illustration of Image Distortions used for Robustness Evaluation}
\label{fig:example_sample}
\vspace{-1.50em}
\end{figure}

Figure~\ref{fig:heatmaps_ssl_accuracy} presents heatmaps of Top-1 and Top-5 linear classification accuracy across CIFAR-100 and Imagenette under five types of distortions. The proposed method consistently achieves the highest accuracy across both datasets and distortion types, outperforming all compared SSL baselines. For instance, under Gaussian noise, our method achieves a Top-1 accuracy of 75.8\% on CIFAR-100 and 82.9\% on Imagenette, significantly surpassing the next best performers. Similarly, in the Top-5 evaluation, our method yields 92.0\% and 92.8\% accuracy under noise on CIFAR-100 and Imagenette, respectively. These results highlight the effectiveness of our approach not only in learning invariant representations but also in preserving semantic discriminability across object classes, making it well-suited for downstream watermarking tasks that demand both robustness and generalization. 
A sample set of original and distorted images used in these experiments is illustrated in Figure~\ref{fig:example_sample}.

\subsection{Robust Zero Watermarking}
\label{sec:wm}

This subsection presents an evaluation of the proposed learning-based zero-watermarking scheme using the invariant features extracted in Module 1. We organize the analysis into three parts. First, we visualize the training dynamics of the reference signature \( C \), including t-SNE projections before and after optimization, and monitor the convergence behavior of the watermark loss over time. Next, we quantify the robustness of the watermarking system across a wide range of image distortions by evaluating bit accuracy under varying transformation intensities. Finally, we compare our method against existing state-of-the-art zero-watermarking and other deep watermarking approaches, highlighting the performance gains achieved by our framework in terms of robustness and stability. While most previous studies have relied on pretrained models to extract features for robust zero-watermarking, we used our own invariant feature model to enhance robustness. Together, these experiments establish the practicality and generalizability of our proposed method.
We evaluated the proposed method using binary watermark messages of varying lengths and report results using a 30-bit watermark to align with common practice in the literature and ensure fair comparison with prior work.

\subsubsection{Visualization of Reference Signature Learning}
\label{sec:c_learning}

\begin{figure}[h]
\centering
\includegraphics[width=0.65\linewidth]{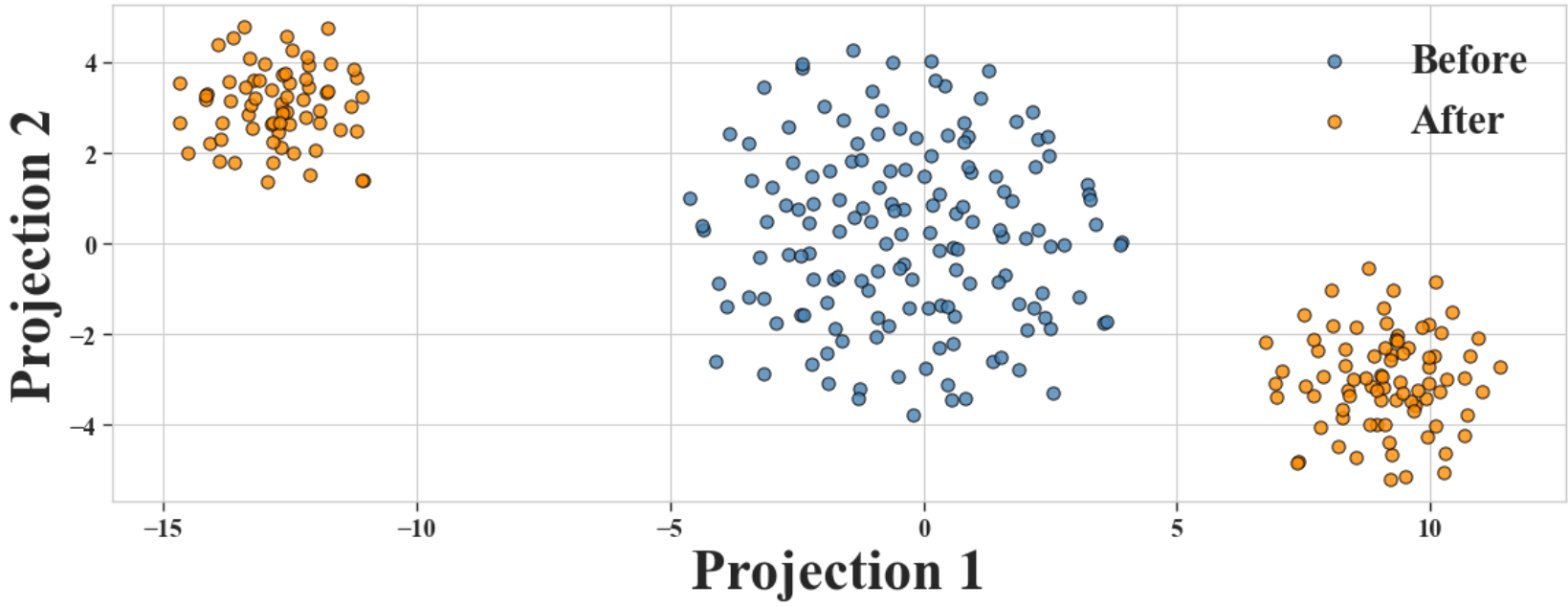}
\vspace{-0.75em}
\caption{ t-SNE Projection of Reference Signatures $C$ Before and After Training.}
\label{fig:c_learning}
\vspace{-0.75em}
\end{figure}

To understand the optimization dynamics of the learned reference signature \( C \), we visualize both its geometric evolution in feature space and the corresponding training loss over time. Figure~\ref{fig:c_learning} presents two complementary views of this process: the structural arrangement of \( C \) via t-SNE projection, and the loss curves from training the watermark for a few representative image-watermark pairs.

As shown in Fig.~\ref{fig:c_learning}, we apply t-distributed Stochastic Neighbor Embedding (t-SNE) to project the high-dimensional reference vectors \( C \in \mathbb{R}^{k \times d} \) into two dimensions, both before and after training. Each point represents the projection of one bit vector \( C_i \) associated with a distorted image. Before training, the projected vectors appear scattered without clear structure, reflecting the random initialization. After optimization, the vectors form distinct and compact clusters in the embedding space. This emergent separation indicates that the learning process successfully aligns the reference signature with the distortion-invariant feature space. Notably, the improved spatial organization mirrors the expected behavior of our theoretical formulation, in which \( C \) is optimized to serve as a directional codebook that maximizes the margin between binary watermark bits. The optimization implicitly identifies a hyperplane such that \( \tilde{F} \cdot C_i \geq 0.5 \) or \( < 0.5 \), depending on the bit label, enabling robust watermark recovery through simple dot-product thresholding.

\begin{figure}[h]
\centering
\vspace{-0.75em}
\includegraphics[width=0.65\linewidth]{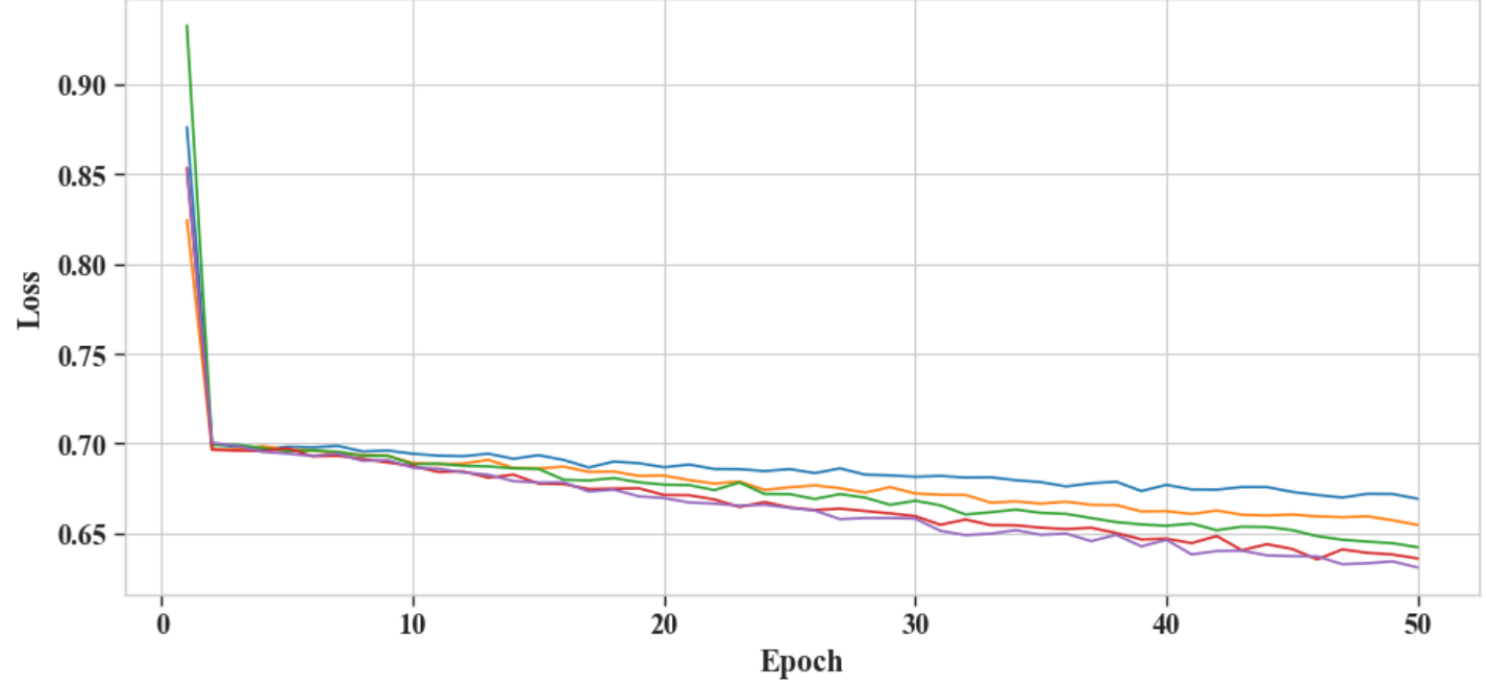}
\vspace{-0.75em}
\caption{Training Loss for Random Watermarking Images.}
\label{fig:c_loss}
\vspace{-1.00em}
\end{figure}

Figure~\ref{fig:c_loss} shows the training loss curves over 50 epochs for five randomly selected images. The consistent and monotonic reduction in loss illustrates effective convergence of the reference optimization process, with the steep early drop reflecting rapid alignment with the underlying feature distribution. Together with the t-SNE results, this confirms that the learned signatures not only generalize across distorted views but are also geometrically structured in a way that supports discriminative watermark decoding. These findings reinforce the strength of learning-based watermark representation as a robust alternative to traditional hand-crafted correlation schemes.

\subsubsection{Distortion Tolerance Range}
\label{sec:tolerance_range}

\begin{figure}[h]
\centering
\vspace{-0.75em}
\includegraphics[width=0.65\linewidth]{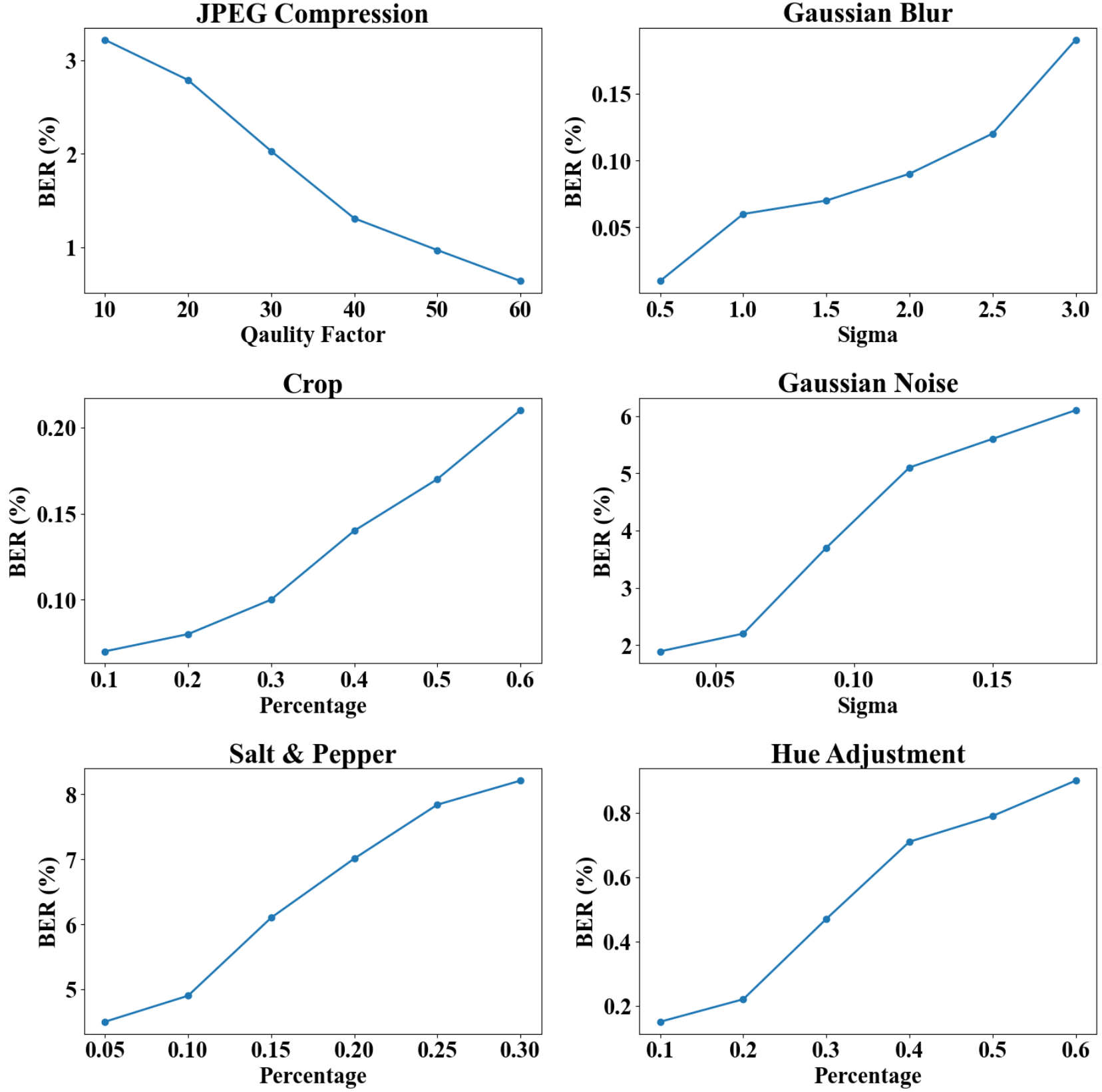}
\caption{BER (\%) under Various Distortion Types.}
\label{fig:tolarance}
\vspace{-0.75em}
\end{figure}

Figure~\ref{fig:tolarance} illustrates the Bit Error Rate (BER) performance of the proposed watermarking system under six common example distortions: JPEG compression, Gaussian blur, cropping, Gaussian noise, salt \& pepper noise, and hue adjustment. BER is computed as the ratio of incorrectly extracted bits to the total number of embedded bits, expressed as a percentage. All results are averaged over 1,000 test images, each containing a 30-bit binary watermark, resulting in a total of 30,000 evaluated bits per distortion setting.

For JPEG compression, BER consistently decreases as the quality factor improves, confirming strong robustness to compression artifacts. Gaussian blur exhibits a gradual BER increase with higher blur radii, while cropping and hue adjustment maintain relatively low BER values even at high distortion intensities. This indicates that the proposed method effectively preserves watermark integrity under spatial and color-shift perturbations. 
Noise-based distortions, especially Gaussian noise and salt \& pepper noise, introduce relatively more significant degradation, with BER rising sharply under higher noise levels. Nonetheless, the absolute BER values remain low across all cases, ranging from 0.05\% to a maximum of 8\%, highlighting the overall robustness of the proposed method.

\subsubsection{Comparison with Existing Watermarking Methods}
\label{sec:wm_comparison}

Table~\ref{tab:framework_comparison} highlights the key distinctions between the proposed framework and existing approaches. Traditional and deep zero-watermarking methods rely on fixed or task-driven features without explicitly enforcing distortion invariance, resulting in limited robustness under diverse corruptions. SSL-based representations provide general semantic invariance but are not optimized to suppress distortion-specific variability that directly affects watermark signatures.

\vspace{-1.5em}
\begin{table}[ht]
\caption{Comparison of watermarking paradigms and invariant representation properties.}
\label{tab:framework_comparison}
\centering
\resizebox{0.95\linewidth}{!}{
\begin{tabular}{|l|c|c|c|c|c|}
\hline
Method & Training Paradigm & Invariance Mechanism & Feature Stability & Generalization to Unseen Distortions & Multi-bit \\
\hline
Traditional Zero-WM \cite{anand2024authenticating}  & Handcrafted & None (fixed design) & Limited, distortion-specific & Weak & Limited \\
Deep Zero-WM \cite{gharib2025robust} & Task-driven CNN & Implicit (data-dependent) & Moderate & Limited & \cmark \\
SSL Feature-based \cite{fernandez2022watermarking} & Pre-trained (task-agnostic) & Augmentation invariance & Semantic but not distortion-stable & Moderate & Limited \\
Deep WM (Embedding) \cite{zhu2018hidden} & End-to-end embedding & Learned robustness (task-specific) & In-domain & Limited (overfits distortions) & \cmark \\
\textbf{InvZW (Ours)} & Invariant representation learning & Adversarial + reconstruction & \textbf{High, distortion-invariant} & \textbf{Strong (unseen distortions)} & \textbf{\cmark} \\
\hline
\end{tabular}}
\end{table}
\vspace{-1.5em}

In contrast, the proposed framework explicitly learns distortion-invariant representations through adversarial alignment between clean and distorted inputs, while a reconstruction constraint preserves semantic structure and prevents feature collapse. This combination enables stable feature representations that generalize beyond the distortions observed during training, providing improved robustness for multi-bit zero-watermarking under diverse real-world conditions.

To quantitatively evaluate the robustness of the proposed watermarking framework, we compare it against several representative models across three categories: (1) deep learning-based watermarking schemes that embed information into the image, (2) feature extractor-based watermarking methods that leverage pre-trained SSL models, and (3) zero-watermarking approaches with deep models. Each comparison focuses on bit-level accuracy or BER under various common distortions. 

For a fair comparison, all methods were evaluated using a fixed watermark length of 30 bits per image. Additionally, watermarking embedding was controlled to ensure consistency across methods. In particular, for embedded watermarking schemes, watermarking were applied such that the resulting marked images maintained a PSNR of approximately 40\,dB. This constraint was applied to ensure consistent visual quality across all methods, enabling a fair evaluation of robustness under similar imperceptibility conditions. Although some existing methods do not adopt this practice, using their original configurations could result in marked images with varying PSNR levels, leading to an imbalanced comparison. By standardizing the distortion budget, we provide a more objective and controlled benchmarking environment across both embedded and zero-watermarking schemes. In contrast, the proposed method and other zero-watermarking baselines do not alter the original image, preserving its fidelity by design.

\begin{table}[t] 
\centering 
\caption{Bit Accuracy (\%) Comparison with Watermarking Models that Embed Information.} 
\vspace{0.5em}
\label{tab:bit_accuracy_vertical} 
\resizebox{0.45\linewidth}{!}{%
\begin{tabular}{|l|c|c|c|}
\hline
\textbf{Distortion} & \textbf{\cite{zhu2018hidden}} & \textbf{\cite{luo2020distortion}} & \textbf{Ours} \\ 
\hline
Gaussian Noise (0.06) & 74.6 & 95.6 & \textbf{97.8} \\ 
Gaussian Noise (0.08) & 67.7 & 93.5 & \textbf{96.7} \\ 
Gaussian Noise (0.10) & 63.2 & 89.5 & \textbf{95.3} \\ \hline
Salt \& Pepper (0.05) & 99.1 & 95.7 & \textbf{95.5} \\ 
Salt \& Pepper (0.10) & 93.1 & 85.0 & \textbf{95.1} \\ 
Salt \& Pepper (0.15) & 83.4 & 77.1 & \textbf{93.9} \\ \hline
Adjust Hue (0.2) & 65.1 & 94.0 & \textbf{99.8} \\ 
Adjust Hue (0.4) & 34.0 & 70.7 & \textbf{99.3} \\ 
Adjust Hue (0.6) & 18.1 & 42.4 & \textbf{99.1} \\ \hline
Adjust Saturation (5.0) & 99.9 & 95.2 & \textbf{99.9} \\ 
Adjust Saturation (10.0) & 99.6 & 93.3 & \textbf{99.4} \\ 
Adjust Saturation (15.0) & 98.5 & 90.1 & \textbf{98.7} \\ \hline
GIF (64) & 87.1 & 97.6 & \textbf{99.5} \\ 
GIF (32) & 76.8 & 95.7 & \textbf{97.7} \\ 
GIF (16) & 65.0 & 91.7 & \textbf{93.9} \\ \hline
Resize Width (0.9) & 99.3 & 99.9 & \textbf{99.9} \\ 
Resize Width (0.7) & 85.3 & 88.4 & \textbf{97.3} \\ 
Resize Width (0.5) & 66.5 & 67.1 & \textbf{94.8} \\ 
\hline
\textbf{Average} & 76.5 & 86.8 & \textbf{96.5} \\ 
\hline
\end{tabular}%
}
\vspace{-0.75em}
\end{table}

Table~\ref{tab:bit_accuracy_vertical} presents bit accuracy results under a wide range of distortions. We compare our method with HIDDEN~\cite{zhu2018hidden}, a widely used encoder-decoder watermarking scheme, and~\cite{luo2020distortion}, which uses adversarial training to improve robustness. Our model consistently achieves the highest accuracy across most transformations, particularly under color-related distortions such as hue and saturation adjustment, where alternative methods degrade significantly. Moreover, our model shows stronger robustness under severe noise (e.g., Gaussian 0.10, Salt \& Pepper 0.15), outperforming others by large margins. The average accuracy improvement of our model demonstrates superior generalization across both geometric and photometric distortions.

\begin{table}[t] 
\centering 
\caption{Bit Accuracy (\%) Comparison with Pretrained Feature-Based Watermarking Models.} 
\vspace{0.5em}
\label{tab:bit_accuracy} 
\resizebox{0.45\linewidth}{!}{%
\begin{tabular}{|l|c|c|c|}
\hline
\textbf{Distortion} & \textbf{\cite{vukotic2020classification}} & \textbf{\cite{fernandez2022watermarking}} & \textbf{Ours} \\ 
\hline
Identity & 1.00 & 1.00 & \textbf{1.00} \\ 
Rotation (25) & 0.27 & 1.00 & \textbf{1.00} \\ 
Crop (0.5) & 1.00 & 1.00 & \textbf{1.00} \\
Crop (0.1) & 0.02 & 0.98 & \textbf{1.00} \\ 
Resize (0.7) & 1.00 & 1.00 & \textbf{1.00} \\ 
Blur (2.0) & 0.25 & 1.00 & \textbf{1.00} \\
JPEG (50) & 0.96 & 0.97 & \textbf{0.99} \\ 
Brightness (2.0) & 0.99 & 0.96 & \textbf{1.00} \\ 
Contrast (2.0) & 1.00 & 1.00 & \textbf{1.00} \\
Hue (0.25) & 1.00 & 1.00 & \textbf{1.00} \\ 
Screenshot & 0.86 & 0.97 & \textbf{0.98} \\ 
\hline
\end{tabular}%
}
\vspace{-0.50em}
\end{table}

Table~\ref{tab:bit_accuracy} compares our framework with two recent methods that utilize pre-trained SSL models for watermark retrieval: Vukotić et al.~\cite{vukotic2020classification}, which uses convolutional networks, and Fernández et al.~\cite{fernandez2022watermarking}, which employs DINO features. Although these approaches demonstrate strong performance under identity-preserving transformations, they often struggle with heavy attacks and geometric alterations such as rotation or crop. In contrast, our model maintains perfect or near-perfect bit accuracy across all comparison distortions. Notably, under non-differentiable transformations like screenshots, our method preserves 98\% bit accuracy, further validating the resilience conferred by our invariant feature learning.

\begin{table}[t] 
\centering 
\caption{BER (\%) Comparison across Zero-Watermarking Models.} 
\vspace{0.5em}
\label{tab:ber} 
\resizebox{0.4\linewidth}{!}{%
\begin{tabular}{|l|c|c|}
\hline
\textbf{Distortion} & \textbf{\cite{he2023shrinkage}} & \textbf{Ours} \\ 
\hline
Gaussian Noise (0.005) & 0.0151 & \textbf{0.0139} \\ 
Gaussian Noise (0.01) & 0.0231 & \textbf{0.0187} \\ 
JPEG Compression (30) & \textbf{0.0267} & 0.0431 \\
JPEG Compression (70) & \textbf{0.0169} & 0.0748 \\ 
Rotation (3) & 0.0473 & \textbf{0.0196} \\ 
Rotation (5) & 0.0622 & \textbf{0.0237} \\
Scaling (0.5) & 0.0182 & \textbf{0.0143} \\ 
Scaling (2.0) & 0.0153 & \textbf{0.0093} \\ 
Median Filter ($3\times 3$) & \textbf{0.0127} & 0.0151 \\
Median Filter ($5\times 5$) & \textbf{0.0170} & 0.0291 \\ 
Gaussian Filter ($3\times 3$) & 0.0075 & \textbf{0.0038} \\ 
Gaussian Filter ($5\times 5$) & 0.0126 & \textbf{0.0089} \\ \hline
\end{tabular}%
}
\vspace{-0.9em}
\end{table}

Table~\ref{tab:ber} reports BER comparisons with a recent deep zero-watermarking method from~\cite{he2023shrinkage}. Across several distortion types, including geometric transformations and Gaussian filtering, our method demonstrates competitive robustness and often achieves lower BER than the baseline. In particular, under rotation and scaling, as well as Gaussian filtering, the proposed framework consistently yields improved performance.

At the same time, the baseline method performs better under certain distortions, notably JPEG compression and median filtering. In particular, under strong JPEG compression (e.g., quality 30 and 70), the proposed method exhibits higher BER, indicating that although compression artifacts are captured by the learned invariant features, they are not explicitly optimized in the same way as handcrafted or distortion-specific designs. 
These differences reflect the distinct design principles of the two approaches. Existing methods rely on fixed or handcrafted feature stability, which can be effective under specific distortions such as compression, while our method focuses on learning invariant representations that generalize across diverse transformations, particularly geometric and filtering distortions.

Overall, the results suggest that the proposed invariant feature framework provides a flexible and generalizable representation, achieving strong performance across a wide range of distortions while exhibiting complementary behavior to existing methods. 

\vspace{-0.75em}
\subsection{Ablation Study}

To assess the importance of the Reconstructor module \( R \), we conduct an ablation study by removing \( R \) from the training pipeline. As discussed in Section~\ref{sec:methods}, \( R \) is introduced to prevent feature collapse where the feature extractor learns to produce constant or near-zero outputs that are trivially invariant but devoid of semantic content. Although such collapsed representations may fool the distortion discriminator, they lack the richness necessary to support watermarking or downstream tasks. The reconstructor acts as a semantic regularizer, ensuring that invariant features must still encode sufficient information to reconstruct the input image. This auxiliary constraint discourages degenerate solutions by penalizing high reconstruction error when semantic content is not preserved.


Table~\ref{tab:ablation_recon} shows a substantial drop in both linear classification accuracy (obtained by training a shallow classifier on top of frozen features) and watermark bit accuracy when the Reconstructor is removed. Specifically, linear accuracy drops from 81.67\% to 63.31\%, and bit accuracy degrades from 96.7\% to 68.1\%. These declines indicate that the learned features lose both discriminative and semantic structure in the absence of \( R \), which directly impacts its ability to support watermark extraction or classification. 
\begin{table}[h]
\centering
\caption{Reconstructor Improves Feature Semantics and Watermark Recovery}
\label{tab:ablation_recon}
\resizebox{0.4\linewidth}{!}{
\begin{tabular}{ccc}
\toprule
\textbf{Reconstructor} & \textbf{Linear Accuracy (↑)} & \textbf{Bit Accuracy (↑)} \\
\midrule
$\times$ & 63.31\% & 68.1\% \\
\checkmark    & \textbf{81.67\%} & \textbf{96.7\%} \\
\bottomrule
\end{tabular}
}
\end{table}

\begin{figure}[h]
    \centering
    \includegraphics[width=0.65\textwidth]{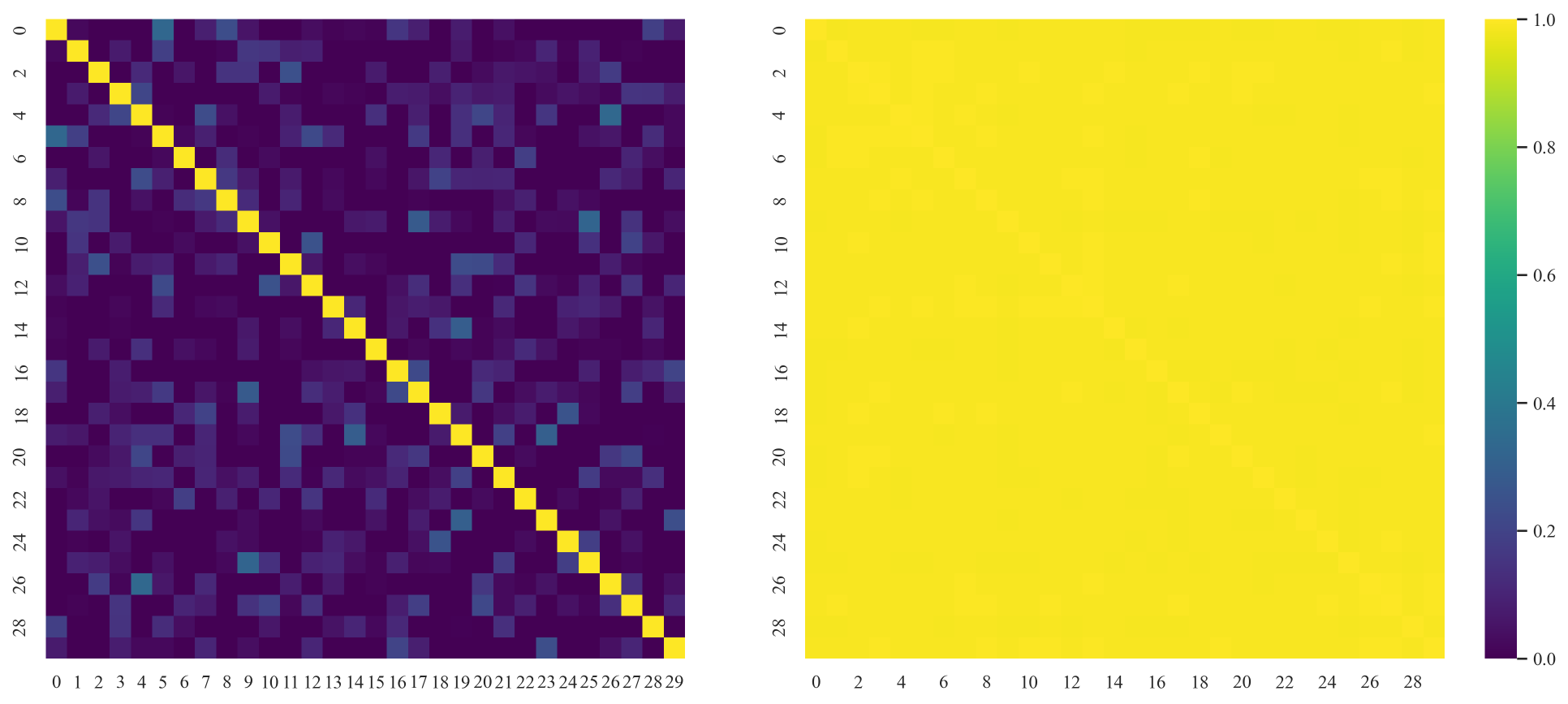}
    \caption{Cosine similarity heatmaps for 30 input images. \textbf{Left:} With Reconstructor, feature vectors are diverse and input-dependent. \textbf{Right:} Without Reconstructor, features collapse to nearly identical outputs.}
    \label{fig:cosine_heatmap}
\end{figure}

To assess feature diversity, we compute the average pairwise cosine similarity between features extracted from different input images across the testing sets.  Fig.~\ref{fig:cosine_heatmap} shows heatmaps for models trained with and without the Reconstructor. 
With the Reconstructor, features exhibit strong diagonal patterns and varied similarities, indicating semantic distinction across inputs. Without it, features become nearly identical, resulting in uniformly high similarity - clear evidence of feature collapse.

\begin{table}[h]
\centering
\caption{Adversarial Training Improves Feature Invariance and Watermark Recovery}
\label{tab:ablation_adv}
\resizebox{0.4\linewidth}{!}{
\begin{tabular}{ccc}
\toprule
\textbf{Adversarial} & \textbf{Linear Accuracy (↑)} & \textbf{Bit Accuracy (↑)} \\
\midrule
$\times$ & 69.45\% & 74.3\% \\
\checkmark & \textbf{81.67\%} & \textbf{96.7\%} \\
\bottomrule
\end{tabular}
}
\end{table}

To evaluate the role of adversarial training, we compare models trained with and without the adversarial objective while keeping other components unchanged. As shown in Table~\ref{tab:ablation_adv}, removing adversarial training leads to a noticeable drop in both linear evaluation accuracy and watermark bit accuracy. This indicates that, without adversarial supervision, the learned features are less robust to distortions and less effective for reliable watermark decoding. In contrast, incorporating adversarial training significantly improves both semantic representation quality and watermark recovery performance, demonstrating its critical role in enforcing distortion-invariant features, preventing feature collapse and ensuring meaningful invariant representations.

In addition, these findings empirically validate the theoretical formulation in Eq.~\ref{eqn: FE}, where the goal is to maximize the mutual information \( I(F, x) \) between the learned features \( F \) and the original image \( x \), while minimizing \( I(F, x') \) with respect to distorted inputs \( x' \). The reconstruction loss enforces the first objective by requiring that \( F \) retain enough signal to recover \( x \), thereby serving as a proxy for maximizing \( I(F, x) \). Without \( R \), this semantic pathway is lost, and the model is free to collapse to trivial invariants. Thus, the reconstructor plays an essential role in balancing invariance and informativeness, acting as a critical regularizer for robust and meaningful representation.

\subsection{Computational Cost and Scalability}
\label{sec:cost}

The proposed framework consists of a ViT-based feature extractor, a discriminator, and a reconstructor. The feature extractor contains approximately 9M parameters, while the discriminator and reconstructor contain approximately 14M and 7M parameters, respectively. Despite the multi-component design, all modules operate on relatively small inputs of size $128 \times 128$, making training computationally practical. In our implementation, the full training process is conducted on a single GPU (Nvidia RTX 4090) and takes approximately 46 hours. The framework can be further scaled by replacing the backbone with larger pre-trained models, without modifying the overall pipeline. 

At inference time, only the feature extractor is required for watermark extraction, while the discriminator and reconstructor are used during training only. This significantly reduces the deployment cost and allows efficient inference comparable to standard feature extraction pipelines.

In terms of storage, the framework requires storing a reference signature tensor of size $128 \times 128 \times 3$ for each image--watermark pair. The storage cost is therefore linear in the number of protected samples and depends on the numerical precision used for the stored features.

%% file: sec/5_conclusion.tex
\section{Conclusion}
This paper presents InvZW, a deep learning-based framework for robust image zero-watermarking built upon invariant feature learning through noise-adversarial training. Unlike traditional or pretrained feature extractor-based methods, our approach explicitly learns distortion-invariant yet semantically meaningful representations via a combination of adversarial alignment and reconstruction constraints. These representations serve as a stable basis for a novel zero-watermarking scheme, where a binary watermark is associated with a reference signature optimized in the invariant feature space.

We demonstrate that the proposed framework not only achieves high robustness against diverse photometric and geometric distortions, but also enables accurate watermark recovery through a novel flexible embedding mechanism. Extensive evaluations show superior performance over representative deep watermarking methods, including encoder-decoder-based, self-supervised, and zero-watermarking baselines.

While this study focuses on a wide range of simulated digital distortions, future work will involve evaluating the proposed framework under real-world perturbations, such as camera-captured compression artifacts, sensor noise, and image degradations from mobile or embedded devices. This will help validate the framework's applicability in operational deployment scenarios. The framework will also be extended in the future to video watermarking and real-time applications, as well as incorporating adaptive optimization strategies for signature learning under dynamic distortion environments. Moreover, future evaluations will explore the method's robustness under more extreme and semantic distortions, such as style transfer and image enhancement, to further assess its generalization in real-world scenarios. The integration of learned invariance and adaptive encoding opens new possibilities for secure, high-fidelity media protection in both static and streaming contexts.